\documentclass[]{style}

\usepackage[toc,page,header]{appendix}
\usepackage{enumitem}
\usepackage{hyperref}

\usepackage{minitoc}
\usepackage{natbib}
\usepackage{CJKutf8}

\usepackage{xargs}  

\usepackage{todonotes}  

\usepackage{multirow}

\usepackage{amsmath}
\usepackage{dsfont}

\usepackage{svg}

\usepackage{mathrsfs}
\usepackage{adjustbox}
\usepackage{multirow}
\usepackage{multicol}
\usepackage{tcolorbox}
\usepackage{changepage}
\usepackage{enumitem}
\usepackage{graphicx}
\usepackage{amssymb}
\usepackage{xcolor}
\usepackage{float}
\usepackage{multirow}
\usepackage{threeparttable}
\usepackage{graphicx}
\usepackage{subcaption}
\usepackage{algorithm}
\usepackage{algpseudocode}
\usepackage{wrapfig}
\usepackage[table]{xcolor}  
\usepackage{colortbl}       
\usepackage{tabularx} 
\usepackage{makecell} 
\usepackage[dvipsnames]{xcolor}

\newcolumntype{g}{>{\columncolor{gray!10}}c} 

\definecolor{catgray}{gray}{0.9}
\definecolor{skyblue}{rgb}{0.53,0.81,0.92} 

\colorlet{skyblue!30}{skyblue!30!white} 

\definecolor{customblue}{RGB}{70,130,180}  

\newtcolorbox{evolbox}[2][]{%
  enhanced,
  colframe=customblue,
  colback=white,
  coltitle=white,
  rounded corners,
  boxrule=1pt,
  titlerule=0pt,
  toptitle=1mm,
  bottomtitle=1mm,
  fonttitle=\bfseries,
  width=#2\textwidth, 
  #1
}

\usepackage{url}

\PassOptionsToPackage{table,xcdraw}{xcolor}
\usepackage{placeins}
\usepackage{pifont}

\definecolor{RowBlue}{HTML}{E9F2FB}
\definecolor{RowRed}{HTML}{F9EAEA}
\definecolor{Top1}{HTML}{FCE3E3} 
\definecolor{Top2}{HTML}{FFF5D1} 
\definecolor{Top3}{HTML}{E1ECF7} 
\definecolor{Sub1}{HTML}{C7DBF2}
\definecolor{Sub2}{HTML}{E4E4E4}

\renewcommand{\emph}[1]{\textit{#1}}

\title{PhysisForcing: Physics Reinforced World Simulator for Robotic Manipulation}

\author[*]{Peiwen Zhang}
\author[*]{Yufan Deng}
\author[*]{Shangkun Sun}
\author{Juncheng Ma}
\author[\dagger]{Duomin Wang}
\author{Jonas Du}
\author{Zilin Pan}
\author{Ye Huang}
\author{Hao Liang}
\author{Songyan Huang}
\author{Ruihua Zhang}
\author[\dagger]{Enze Xie}
\author{Ming-Yu Liu}
\author[\ddagger]{Daquan Zhou}

\affiliation{Peking University}
\affiliation{NVIDIA}

\contribution[*]{Equal Contribution}
\contribution[\dagger]{Co-Project Lead}
\contribution[\ddagger]{Corresponding Author}

\definecolor{PekingRed}{RGB}{178,31,45}

\abstract{
Video generation models have emerged as a promising paradigm for embodied world simulation. However, both general-domain video generators and robot-specific data fine-tuned models can still produce physically implausible manipulations, including discontinuous motion trajectories and inconsistent robot-object interactions, which limits their reliability as world simulators. Through extensive experiments, we find that such physical instability mainly arises from two factors: deformation of moving objects and implausible spatio-temporal correlations among interacting entities, particularly during contact. Building on this observation, we propose \textbf{PhysisForcing}, a scalable training framework that strengthens physical consistency by focusing supervision on physics-informative regions through joint optimization of pixel-level and semantic-level features. The framework consists of a pixel-level trajectory alignment loss, which supervises DiT features using reference point trajectories, and a semantic-level relational alignment loss, which aligns DiT features with inter-region relations extracted from a frozen video understanding encoder. Extensive experiments on R-Bench, PAI-Bench, and EZS-Bench show that PhysisForcing consistently improves embodied video generation over strong baselines, improving the Wan2.2-I2V-A14B and Cosmos3-Nano base models on R-Bench by 22.3\% and 9.2\% (7.1\% and 3.7\% over vanilla finetuning), with the Cosmos3-Nano variant attaining the best overall score. Beyond generation, as a world model under the WorldArena action-planner protocol it raises the closed-loop success rate from 16.0\% to 24.0\% and further improves downstream policy success, indicating that physically aligned video models yield stronger representations for robotic manipulation. Code and more results are available at \url{https://dagroup-pku.github.io/PhysisForcing.github.io/}.
}

\begin{document}
\maketitle

\begin{figure*}[!h]
\centering
\includegraphics[width=\textwidth]{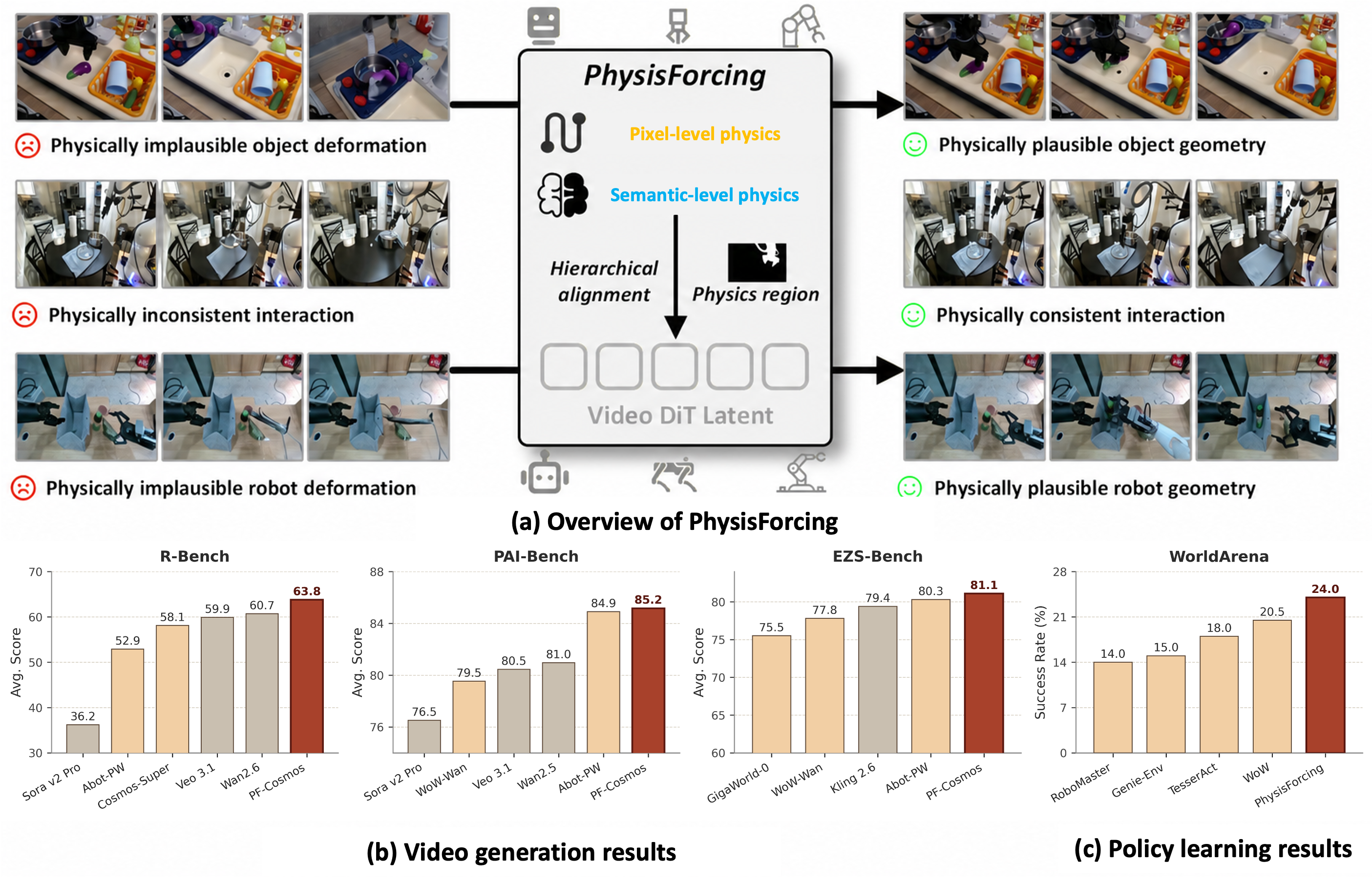}
\caption{\textbf{Impacts of PhysisForcing on video generation and robotics policy learning.} Our method introduces hierarchical physics alignment during video generation training, enforcing both pixel-level motion consistency and semantic-level relational coherence. This dual-level supervision enables generation of robotic manipulation videos that are not only visually realistic but also physically plausible and useful for downstream robot action modeling. In the generation benchmarks (R-Bench, PAI-Bench, EZS-Bench), \textbf{PF-Cosmos} denotes Cosmos3-Nano trained with PhysisForcing; the WorldArena results use the Wan2.2-TI2V-5B world model trained with PhysisForcing.}
\label{fig:teaser}
\end{figure*}

\section{Introduction}

Video generation models have shown strong potential as world simulators for embodied intelligence~\cite{openai2024sora,GoogleDeepMind2025Veo3,wan2025wan,kong2024hunyuanvideo,yang2024cogvideox}, providing scalable visual futures that can be used for data generation, world simulation, and downstream applications such as policy learning~\cite{bruce2024genie,o2024open,nvidia2026worldsimulationvideofoundation,ye2026worldactionmodelszeroshot,li2025unified,liang2025video, lingbot-va2026}. However, embodied world simulation requires more than photorealistic videos. The generated dynamics must also be physically plausible, especially in contact-rich manipulation.

This requirement remains difficult for current video generators. In contact-rich manipulation, physical violations often appear as local dynamic errors, such as discontinuous gripper trajectories, object penetration, or anti-gravity motion, and as global relational errors, such as a pushed object remaining static or a grasped object drifting away. These errors directly undermine the reliability of the generated video as a visual world simulator, since the predicted frames no longer represent a valid consequence of the robot action.

Existing approaches provide only partial solutions. General video models rely on large-scale visual pre-training but lack sufficient exposure to embodied contact dynamics~\cite{GoogleDeepMind2025Veo3,wan2025wan,kong2024hunyuanvideo,yang2024cogvideox}. Robot-oriented video or world models improve task relevance with manipulation data~\cite{jang2025dreamgen,feng2025vidar,cheang2024gr,du2023learning, chen2025lvp}, yet are usually trained with reconstruction objectives that treat physically critical regions and background pixels uniformly. Recent physics-aware methods introduce geometric cues such as depth~\cite{yang2024depth, video_depth_anything}, tracking~\cite{karaev2023cotracker,karaev24cotracker3,harley2025alltracker}, and 3D structure~\cite{zhen2025tesseract,shang2025roboscape,zhou2024robodreamer,guo2025ctrl}, or apply preference and reward alignment to suppress implausible generations~\cite{ouyang2022training,rafailov2024direct,ali2025abot,zhang2026mind}. However, geometric constraints mainly capture local motion consistency, while preference feedback is often sparse and weakly localized. As a result, existing methods still lack a unified training-time mechanism for aligning both local dynamics and global interaction outcomes.

This raises a key question: \textit{how can we inject physical supervision into video generation in a way that is both hierarchical and region-focused?} We observe that physical plausibility in manipulation videos is naturally hierarchical. At the pixel level, local motion should satisfy trajectory continuity, depth consistency, and contact-compatible displacement. At the semantic level, object relations should evolve according to the interaction semantics: a pushed object should move away, a grasped object should remain coupled with the gripper, and a placed object should rest on the support surface. Meanwhile, physical evidence is highly localized around manipulators, objects, contacts, and moving regions. Applying supervision uniformly over all pixels dilutes these signals and weakens alignment.

Based on this observation, we propose \textbf{PhysisForcing}, a region-focused hierarchical physics alignment framework for video generation. Instead of supervising all pixels uniformly, PhysisForcing first identifies physics-informative regions, including manipulators, manipulated objects, contact areas, and moving parts, and then applies two complementary alignment losses to these regions during backbone training. Specifically, the \textit{pixel-level physics alignment} module uses point tracking~\cite{karaev24cotracker3} to supervise per-point trajectories of the DiT feature, so that local motion stays continuous and contact-compatible. The \textit{semantic-level physics alignment} module instead aligns the pairwise token-similarity matrix of the DiT feature with that of a frozen video understanding encoder~\cite{bardes2024revisiting} on the same tokens, encouraging globally consistent robot-object interactions. In this way, PhysisForcing improves physical plausibility through hierarchical supervision on regions most relevant to robot-object interactions.

Extensive experiments on R-Bench, PAI-Bench, and EZS-Bench demonstrate the effectiveness of PhysisForcing for embodied video generation. Across the Wan2.2-I2V-A14B and Cosmos3-Nano backbones, our method improves physical plausibility scores by 22.3\% and 9.2\% over the base models, and by 7.1\% and 3.7\% over vanilla finetuning, with the Cosmos3-Nano variant attaining the best overall R-Bench score among all evaluated models. Beyond generation, under the WorldArena action-planner protocol~\cite{shang2026worldarena} PhysisForcing lifts the closed-loop success rate from 16.0\% to 24.0\%, surpassing strong world-model planners, and it also improves downstream policy success when used as the video backbone of a world action model~\cite{yuan2026fastwam}, indicating that physically aligned video models provide stronger representations for embodied intelligence.

In summary, our contributions are:
\begin{itemize}
    \item We formulate physical plausibility in robotic video generation as a hierarchical and region-focused alignment problem.
    \item We propose \textbf{PhysisForcing}, a training-time framework that focuses physical supervision on interaction-critical regions and aligns generation at both the pixel-level motion and the semantic-level inter-region relations.
    
    \item We demonstrate consistent gains on R-Bench, PAI-Bench, and EZS-Bench across diffusion video backbones of different scales and families, including Wan2.2-I2V-A14B and Cosmos3-Nano, with ablations validating the importance of hierarchical physical supervision. We further show that PhysisForcing benefits embodied decision making, improving world-model action planning on WorldArena and downstream policy success when used as the video backbone of a world action model.
\end{itemize}
\section{Related Work}

\subsection{Embodied Video Generation and World Models}

Video generation models have emerged as a promising paradigm for embodied intelligence, offering scalable solutions to the data scarcity challenge in robotics~\cite{openai2024sora,wan2025wan,kong2024hunyuanvideo,yang2024cogvideox,GoogleDeepMind2025Veo3}. General-domain models like Sora~\cite{openai2024sora} and Wan~\cite{wan2025wan} demonstrate impressive visual fidelity but often struggle with embodied manipulation tasks due to limited exposure to robot-object interaction dynamics. To address this, recent work has developed robotics-specific world models. Cosmos~\cite{nvidia2026worldsimulationvideofoundation}, DreamGen~\cite{jang2025dreamgen}, and GR-2~\cite{cheang2024gr} focus on action-conditioned video generation for robotic scenarios, while Vidar~\cite{feng2025vidar}, UnifoLM-WMA-0~\cite{unifolm-wma-0}, and Unified Video Action Model~\cite{li2025unified} emphasize world action modeling. Video generators have also been explored as robot policies~\cite{liang2025video,hu2024video}. Gen2Act~\cite{bharadhwaj2024gen2act} generates human videos in novel scenarios to enable generalizable manipulation. Genie~\cite{bruce2024genie} and Genie Envisioner~\cite{liao2025genie} explore interactive world models. However, these approaches primarily optimize visual quality or action controllability, without explicitly enforcing physical constraints during training. Consequently, they remain susceptible to physically implausible manipulation dynamics, including object penetration, trajectory discontinuities, and interaction outcomes that violate basic physical causality, which limits their utility as world simulators.

\subsection{Physics-aware World Simulator}

Recognizing the limitations of purely visual objectives, recent work has begun incorporating physical constraints into video generation. These approaches can be broadly categorized into three paradigms. \textit{Geometry-based methods} enforce pixel-level physical consistency through depth prediction~\cite{yang2024depth,video_depth_anything}, keypoint tracking~\cite{karaev2023cotracker,karaev24cotracker3,harley2025alltracker}, or 3D scene reconstruction~\cite{zhen2025tesseract}. RoboScape~\cite{shang2025roboscape} jointly learns temporal depth prediction and keypoint dynamics to improve geometric consistency and motion modeling. RoboDreamer~\cite{zhou2024robodreamer} learns compositional world models for robot imagination. CTRL-World~\cite{guo2025ctrl} proposes a controllable generative world model. While effective at capturing local spatial structure, these methods do not explicitly model semantic-level interactions or relational outcomes. \textit{Preference-based methods} apply post-training alignment~\cite{ouyang2022training,rafailov2024direct} to suppress physically implausible generations. ABot-PhysWorld~\cite{ali2025abot} employs DPO-based training with physics-aware discriminators to reduce violations like object penetration and anti-gravity motion. MIND-V~\cite{zhang2026mind} uses GRPO with a Physical Foresight Coherence reward to align generated dynamics with physical laws. While these approaches successfully reduce certain failure modes, they operate post-hoc—correcting rather than preventing physical violations—and may sacrifice visual quality in the alignment process. \textit{Simulator-based methods} leverage physics engines~\cite{wang2023robogen,yang2025physics} or test-time selection to ensure physical validity, but incur significant computational overhead and limit scalability.

Existing physics-aware video models often provide supervision at a single level or over the entire frame. However, contact-rich manipulation calls for two design principles. First, physical consistency should be hierarchical: pixel-level motion constraints govern per-point trajectories and contacts, while semantic-level relational cues capture inter-region interactions. Second, physical supervision should be localized: the most informative evidence lies around manipulators, contact interfaces, and moving parts, rather than static background regions. PhysisForcing follows these principles by introducing a region-focused hierarchical physics objective for training video generators.
\section{Method}

\begin{figure*}[!h]
\centering
\includegraphics[width=\textwidth]{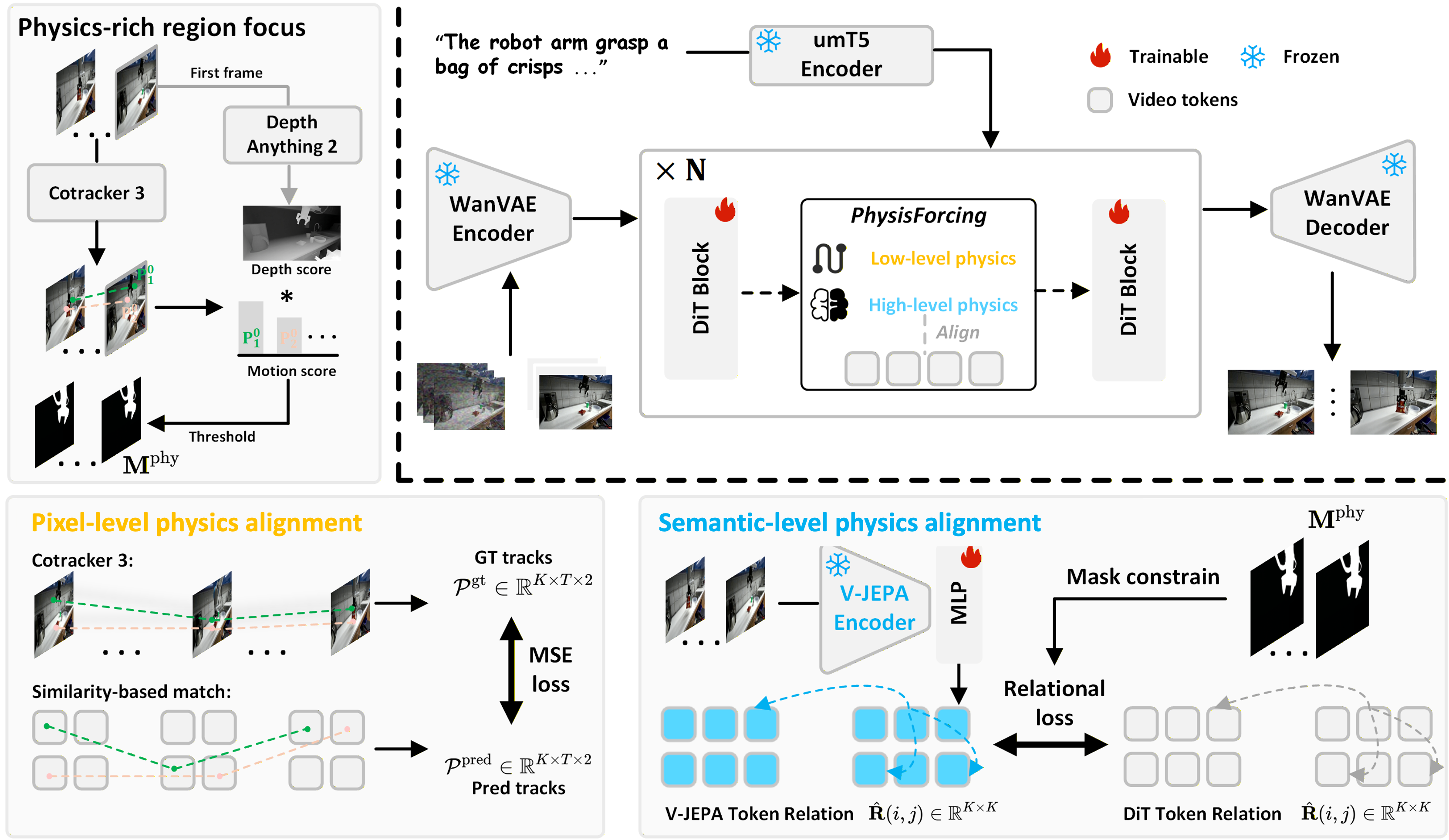}
\caption{\textbf{Overall architecture of PhysisForcing} Our method introduces hierarchical physics alignment during video generation training, enforcing both pixel-level motion consistency (trajectory continuity, contact dynamics) and semantic-level relational consistency (spatio-temporal relations among the robot, object, and scene).}
\label{fig:method_overview}
\end{figure*}

As shown in Figure~\ref{fig:method_overview}, PhysisForcing injects physics supervision into
video generation through a region-focused hierarchical alignment framework. We first identify physics-informative regions where robot-object interactions occur, and then apply
two complementary training signals: pixel-level trajectory alignment for local motion consistency and
semantic-level relational alignment for interaction outcome consistency.

\subsection{Physics-informative Region Extraction}

Embodied video generation errors often occur in contact-rich regions, where robot-object interactions
manifest as large local motion in foreground areas. Therefore, PhysisForcing first identifies such
physics-informative regions from the reference video before applying physics supervision.

Given an input video $V\in\mathbb{R}^{T\times C\times H\times W}$, an off-the-shelf point
tracker~\cite{karaev24cotracker3} is used to obtain dense temporal trajectories
$\mathcal{P}=\{\mathbf{p}_i^{1:T}\}_{i=1}^{N}$, where $N=H\times W$ and
$\mathbf{p}_i^t\in\mathbb{R}^2$ denotes the tracked 2D location of point $i$ at frame $t$:
\begin{equation}
a_i=\sum_{t=1}^{T-1}\left\|\mathbf{p}_i^{t+1}-\mathbf{p}_i^{t}\right\|_2 ,
\end{equation}
where a larger $a_i$ indicates stronger local motion. However, motion magnitude alone may also
highlight background jitter or irrelevant moving regions. Since robot-object interactions are usually
distributed in foreground areas, we introduce a depth-aware foreground weight based on the first
frame. Let $D_0\in\mathbb{R}^{H\times W}$ denote the estimated depth map of the first frame. For each
query point $\mathbf{p}_i^0$, we compute its foreground weight $r_i$ and physics-informative score $q_i$ as:
\begin{equation}
r_i=\frac{1}{D_0(\mathbf{p}_i^0)+\epsilon},
\quad
q_i=a_i\cdot r_i ,
\end{equation}
where $\epsilon$ is a small constant for numerical stability. A larger $q_i$ indicates a trajectory with
both strong local motion and high foreground relevance. We use the mean score as an adaptive threshold to obtain the trajectory-level motion mask:
\begin{equation}
\mathbf{M}^{\mathrm{phy}}_i=\mathbb{I}\left(q_i \geq \frac{1}{N}\sum_{j=1}^{N}q_j\right).
\end{equation}
Finally, we project the selected trajectories onto each frame to form a spatiotemporal physics mask
$\mathbf{M}^{\mathrm{phy}}\in\{0,1\}^{T\times H\times W}$. Specifically, we initialize $\mathbf{M}^{\mathrm{phy}}$ as zeros and set
the pixels visited by selected trajectories to one:
\begin{equation}
\mathbf{M}^{\mathrm{phy}}_t\left(\left\lfloor \mathbf{p}_i^t \right\rceil\right)=1,
\quad \text{if } \mathbf{M}^{\mathrm{phy}}_i=1,\quad t=1,\dots,T .
\end{equation}
Here, $\lfloor \cdot \rceil$ denotes rounding to the nearest pixel.

This mask localizes physics-informative regions where robot-object interactions are likely to occur,
providing spatial guidance for semantic-level physics supervision and pixel-level trajectory alignment.

\subsection{Pixel-level Physics Alignment}

While the physics mask identifies interaction-relevant regions, it does not explicitly constrain
fine-grained point motion. We therefore introduce a pixel-level trajectory alignment loss that
operates at the level of individual tracked points, directly enforcing per-point trajectory
continuity on the manipulator and the manipulated object.

Given the predicted video representation inside the denoising transformer, we extract the hidden
feature $\mathbf{H}^{l}$ from an intermediate DiT block (i.e., $l$ is a middle layer of the DiT, which
we empirically find to carry the most informative motion and structure cues) and refine it with a
lightweight MLP $\phi(\cdot)$, obtaining the DiT feature $\phi(\mathbf{H}^{l})$. Then the refined DiT feature is reshaped into frame-wise feature maps $\hat{\mathbf{F}} \in \mathbb{R}^{T\times C\times H\times W}$. Since all trajectories are queried
from the first frame, we use the first-frame feature as the query feature and the features of the
remaining frames as key features:
\begin{equation}
\mathbf{Q}
=
\hat{\mathbf{F}}_{0},
\quad
\mathbf{K}_{t}
=
\hat{\mathbf{F}}_{t},
\quad
t=1,\dots,T-1 .
\end{equation}
For each query point $\mathbf{p}_i^0$, we sample its query feature $\mathbf{Q}(\mathbf{p}_i^0)$ and
compare it with every spatial location in frame $t$. This gives a similarity map
$\mathbf{s}_i^t\in\mathbb{R}^{H\times W}$:
\begin{equation}
\mathbf{s}_i^t(\mathbf{x})=
\frac{
\mathbf{Q}(\mathbf{p}_i^0)^\top \mathbf{K}_t(\mathbf{x})
}{
\sqrt{C}
},
\quad \mathbf{x}\in\Omega ,
\end{equation}
where $\Omega$ denotes the spatial grid of size $H\times W$. We then normalize the similarity map over the spatial dimension and compute the predicted point
location by coordinate expectation:
\begin{equation}
\hat{\mathbf{p}}_i^t
=
\sum_{\mathbf{x}\in\Omega}
\operatorname{Softmax}_{\mathbf{x}}\left(\mathbf{s}_i^t(\mathbf{x})\right)\mathbf{x},
\end{equation}
where $\hat{\mathbf{p}}_i^t\in\mathbb{R}^2$ is the predicted location of point $i$ at frame $t$. Finally, the predicted trajectories are supervised by the reference trajectories extracted from the
reference video using CoTracker3~\cite{karaev24cotracker3}. Let
$\mathcal{P}_{\mathrm{pred}}=\{\mathbf{p}_{i,\mathrm{pred}}^t\}_{i,t}$ denote the trajectories inferred from the predicted
video, and let $\mathcal{P}_{\mathrm{gt}}=\{\mathbf{p}_{i,\mathrm{gt}}^t\}_{i,t}$ denote the corresponding
reference trajectories. We compute their coordinate discrepancy with a masked mean squared error:
\begin{equation}
\mathcal{L}^{\mathrm{phy}}_{\mathrm{pix}}
=
\frac{1}{|\mathbf{M}^{\mathrm{phy}}|}
\left\|
\mathbf{M}^{\mathrm{phy}}\odot
\left(
\mathcal{P}_{\mathrm{pred}}-\mathcal{P}_{\mathrm{gt}}
\right)
\right\|_2^2 .
\end{equation}
where $\mathbf{M}^{\mathrm{phy}}$ is the physics-informative mask obtained in Sec.~3.1, which restricts trajectory
supervision to interaction-relevant regions.

\subsection{Semantic-level Physics Alignment}

Pixel-level trajectory alignment constrains point-wise motion, but manipulation plausibility also
depends on how different regions relate to each other as the interaction unfolds. For instance,
the gripper and the grasped object should stay tightly coupled, and a pushed object should move
jointly with the contact region. Such inter-region coupling is naturally captured by the pairwise
token similarities of frozen self-supervised video understanding
encoders~\cite{bardes2024revisiting}. We therefore use such an encoder as a measurement space and
align the token-to-token similarity matrix of physics-informative tokens between the DiT side and
the encoder side, so that the encoder's relational structure is transferred into the DiT~\cite{zhang2025videorepa, simeoni2025dinov3}.

Given the input video $\mathcal{V}$, we first extract its target representation with a frozen video
understanding encoder. Meanwhile, we take the hidden feature $\mathbf{H}^{l}$ from the same intermediate (middle-layer) DiT
block and project it into the same representation space with a lightweight MLP $\psi(\cdot)$. The
projected DiT feature is then resized to match the spatio-temporal token layout of the encoder feature:
\begin{equation}
\mathbf{F}^{u}
=
\Phi_{u}(\mathcal{V}),
\quad
\hat{\mathbf{F}}^{u}
=
\operatorname{Resize}\left(\psi(\mathbf{H}^{l})\right),
\end{equation}
where $\Phi_{u}(\cdot)$ denotes the video understanding encoder, and
$\hat{\mathbf{F}}^{u}$ is dimensionally aligned with $\mathbf{F}^{u}$ by interpolation and padding.

We resize the physics-informative mask to the common token resolution and use it to select
spatio-temporal tokens from both representations:
\begin{equation}
\hat{\mathbf{F}}^{\mathcal{M}}
=
\{\hat{\mathbf{F}}^{u}_{t,n}\mid (t,n)\in\mathcal{M}\}
\in\mathbb{R}^{K\times C},
\quad
\mathbf{F}^{\mathcal{M}}
=
\{\mathbf{F}^{u}_{t,n}\mid (t,n)\in\mathcal{M}\}
\in\mathbb{R}^{K\times C},
\end{equation}
where $\mathcal{M}$ is the mask-induced token index set, $K$ is the number of selected tokens, and
$C$ is the aligned feature dimension.

For $i,j\in\{1,\ldots,K\}$, we compute the DiT-side and encoder-side relational matrices as:
\begin{equation}
\hat{\mathbf{R}}(i,j)
=
\frac{
\hat{\mathbf{F}}^{\mathcal{M}}_{i}\cdot \hat{\mathbf{F}}^{\mathcal{M}}_{j}
}{
\left\|\hat{\mathbf{F}}^{\mathcal{M}}_{i}\right\|_2
\left\|\hat{\mathbf{F}}^{\mathcal{M}}_{j}\right\|_2
},
\quad
\mathbf{R}(i,j)
=
\frac{
\mathbf{F}^{\mathcal{M}}_{i}\cdot \mathbf{F}^{\mathcal{M}}_{j}
}{
\left\|\mathbf{F}^{\mathcal{M}}_{i}\right\|
\left\|\mathbf{F}^{\mathcal{M}}_{j}\right\|
}.
\end{equation}
Here, $\hat{\mathbf{R}},\mathbf{R}\in\mathbb{R}^{K\times K}$ capture pairwise spatio-temporal relations
among the selected physics-informative tokens. Finally, the semantic-level physics alignment loss is defined as:
\begin{equation}
\mathcal{L}^{\mathrm{phy}}_{\mathrm{sem}}
=
\frac{1}{K^2}
\sum_{i=1}^{K}
\sum_{j=1}^{K}
\left|
\hat{\mathbf{R}}(i,j)-\mathbf{R}(i,j)
\right|.
\end{equation}

\subsection{Training and Inference}

PhysisForcing is applied during the fine-tuning of a pre-trained DiT-based video generation
backbone~\cite{wan2025wan}. We impose both pixel-level trajectory alignment and semantic-level relational alignment on the
intermediate (middle-layer) feature of the DiT, so that the supervision directly regularizes the
representation used for video prediction; we ablate this layer choice in
Sec.~\ref{sec:ablation} (Table~\ref{tab:ablation_layer}). The overall training objective is:

\begin{equation}
\mathcal{L}
=
\mathcal{L}_{\mathrm{FM}}
+
\lambda_{\mathrm{pix}}\mathcal{L}^{\mathrm{phy}}_{\mathrm{pix}}
+
\lambda_{\mathrm{sem}}\mathcal{L}^{\mathrm{phy}}_{\mathrm{sem}},
\end{equation}
where $\mathcal{L}_{\mathrm{FM}}$ is the standard flow matching loss, and
$\lambda_{\mathrm{pix}},\lambda_{\mathrm{sem}}$ balance the two physics losses. All auxiliary models are used
only during training and are discarded at inference. Thus, PhysisForcing introduces no extra inference cost.
\section{Experiments}

\subsection{Experimental Setup}

\paragraph{Training data.}
We train on a filtered subset of the large-scale RoVid-X dataset~\cite{deng2026rethinking}: from its
4M robotic video clips spanning diverse embodiments, tasks, and environments, stricter motion-score,
task-level de-duplication, and clip-text alignment filtering yields about 500K high-quality clips.

\paragraph{Implementation details.} We apply PhysisForcing to three backbones spanning different
scales and families: Wan2.2-I2V-A14B, Wan2.2-TI2V-5B, and Cosmos3-Nano. For the two Wan backbones,
input videos are resized to $640\times480$ with a maximum length of 81 frames, and both are trained
for 20K steps with a global batch size of 128 using the AdamW optimizer with a learning rate of
$1\times10^{-5}$; for Wan2.2-I2V-A14B, we initialize from its high-noise expert and adapt it to
later denoising stages during training. To test the generality of PhysisForcing across model
families and higher-fidelity regimes, we additionally train the Cosmos3-Nano backbone with LoRA,
following its official image-to-video post-training setting, directly at $720$p resolution with up
to $189$ frames. More details can be found in Appendix~\ref{appx:impl}. For brevity, we refer to the
PhysisForcing-trained Cosmos3-Nano and Wan2.2-I2V-A14B as \textbf{PF-Cosmos} and \textbf{PF-Wan},
respectively; in the policy-learning experiments, where only the Wan2.2-TI2V-5B world model is
involved, we simply write \mbox{+\,PhysisForcing}.

\paragraph{Benchmarks.} We evaluate on three embodied video generation benchmarks (details in
Appendix~\ref{appx:benchmarks}): R-Bench~\cite{deng2026rethinking} (650 image-text pairs across
task-oriented and embodiment-specific dimensions), the \emph{robot domain} of the
PAI-Bench~\cite{zhou2025paibench} generation track (PAI-Bench-G, 174 real-world image-prompt
pairs), and EZS-Bench~\cite{ali2025abot}, a training-independent zero-shot benchmark of 196 unseen
robot-task-scene combinations probing out-of-distribution generalization.

\paragraph{Baselines.} We compare with representative models from three families:
general-domain open-source video models (e.g., HunyuanVideo~\cite{wu2025hunyuanvideo}, LTX-Video~\cite{HaCohen2024LTXVideo}, Wan~\cite{wan2025wan}), commercial
models (e.g., Wan2.6~\cite{wan2025wan}, Seedance~\cite{chen2025seedance}, Hailuo~\cite{hailuo2024}, Veo~\cite{GoogleDeepMind2025Veo3}, Kling~\cite{kling2025}, Sora~\cite{openai2025sora2}), and robotics-specific models (e.g., Cosmos~\cite{nvidia2026worldsimulationvideofoundation},
DreamGen~\cite{jang2025dreamgen}, Vidar~\cite{feng2025vidar}, UnifoLM-WMA-0~\cite{unifolm-wma-0}, WoW~\cite{chi2025wow}, Abot-PhysWorld~\cite{ali2025abot}).

\subsection{Evaluation results on Embodied Video Generation}

\begin{table*}[t]
\centering
\renewcommand{\arraystretch}{1.15}
\setlength{\tabcolsep}{3.8pt}
\caption{\textbf{R-Bench quantitative results.} Evaluations across task-oriented and embodiment-specific dimensions for models from open-source, commercial, and robotics-specific families. Per-column rankings are highlighted in \colorbox{Top1}{\strut 1st}, \colorbox{Top2}{\strut 2nd}, and \colorbox{Top3}{\strut 3rd}.}
\resizebox{\textwidth}{!}{%
\begin{tabular}{
l c|
>{\centering\arraybackslash}p{2.1cm} 
>{\centering\arraybackslash}p{2.1cm} 
>{\centering\arraybackslash}p{2.1cm} 
>{\centering\arraybackslash}p{2.1cm} 
>{\centering\arraybackslash}p{2.1cm}|
>{\centering\arraybackslash}p{2.0cm} 
>{\centering\arraybackslash}p{2.0cm} 
>{\centering\arraybackslash}p{2.0cm} 
>{\centering\arraybackslash}p{2.0cm}  
}
\toprule
\textbf{Models} & \textbf{Avg.} &
\multicolumn{5}{c|}{\textbf{Tasks}} &
\multicolumn{4}{c}{\textbf{Embodiments}} \\
 &  &
\textbf{Manipulation} & \textbf{Spatial} & \textbf{Multi-entity} & \textbf{Long-horizon} & \textbf{Reasoning} &
\textbf{Single arm} & \textbf{Dual arm} & \textbf{Quadruped} & \textbf{Humanoid} \\
\midrule
\rowcolor{RowBlue}
\multicolumn{11}{l}{\textbf{\textit{Open-source}}} \\

HunyuanVideo 1.5~\cite{wu2025hunyuanvideo} & 46.0 &
44.2 & 31.6 & 31.2 & 43.8 & 36.4 &
51.3 & 52.6 & 63.4 & 59.5 \\

LongCat-Video~\cite{meituanlongcatteam2025longcatvideotechnicalreport} & 43.7 &
37.2 & 31.0 & 22.0 & 38.4 & 18.6 &
58.6 & 57.6 & 68.1 & 62.1 \\

Wan2.1-14B~\cite{wan2025wan} & 39.9 &
34.4 & 26.8 & 28.2 & 33.5 & 20.5 &
46.4 & 49.7 & 59.5 & 59.9 \\

LTX-2~\cite{hacohen2026ltx} & 38.1 &
28.4 & 30.4 & 23.3 & 38.6 & 16.4 &
45.3 & 42.4 & 62.2 & 55.5 \\

Wan2.2-TI2V-5B~\cite{wan2025wan} & 38.0 &
33.1 & 31.3 & 14.2 & 31.8 & 23.4 &
43.6 & 44.8 & 59.0 & 60.7 \\

SkyReels~\cite{chen2025skyreelsv2infinitelengthfilmgenerative} & 36.1 &
20.3 & 27.6 & 20.3 & 25.4 & 23.4 &
50.7 & 47.7 & 58.6 & 50.9 \\

LTX-Video~\cite{HaCohen2024LTXVideo} & 34.4 &
30.2 & 17.6 & 21.0 & 28.0 & 24.1 &
44.0 & 45.6 & 52.6 & 46.4 \\

FramePack~\cite{zhang2025framepackv1} & 33.9 &
20.6 & 25.8 & 17.3 & 16.9 & 17.0 &
44.0 & 46.4 & 62.6 & 54.8 \\

HunyuanVideo~\cite{kong2024hunyuanvideo} & 30.3 &
17.7 & 18.0 & 10.8 & 14.7 & 3.5 &
45.4 & 48.0 & 62.5 & 52.4 \\

CogVideoX\_5B~\cite{yang2024cogvideox} & 25.6 &
11.6 & 11.2 & 9.8 & 21.2 & 7.9 &
33.8 & 38.5 & 46.5 & 49.6 \\

\midrule
\rowcolor{RowBlue}
\multicolumn{11}{l}{\textbf{\textit{Commercial}}} \\

Wan2.6~\cite{wan2025wan} & 60.7 &
54.6 & 65.6 & 47.9 & 51.4 & \cellcolor{Top1}53.1 &
66.6 & \cellcolor{Top3}68.1 & 72.3 & 66.7 \\

Veo 3.1~\cite{GoogleDeepMind2025Veo3} & 59.9 &
54.1 & 47.4 & \cellcolor{Top1}53.4 & \cellcolor{Top3}59.2 & 46.7 &
67.0 & 66.6 & \cellcolor{Top1}74.3 & \cellcolor{Top1}70.4 \\

Seedance 1.5 Pro~\cite{chen2025seedance} & 58.4 &
\cellcolor{Top3}57.7 & 49.5 & 48.4 & 57.0 & 47.0 &
64.8 & 64.1 & 68.0 & \cellcolor{Top2}69.2 \\

Wan2.5~\cite{wan2025wan} & 57.0 &
52.7 & 57.6 & 40.2 & 49.6 & 43.7 &
\cellcolor{Top3}68.0 & 63.4 & 72.6 & 65.4 \\

Hailuo v2~\cite{hailuo2024} & 56.5 &
56.0 & 63.7 & 38.6 & 54.5 & 47.4 &
59.4 & 61.1 & 64.0 & 63.5 \\

Veo 3~\cite{GoogleDeepMind2025Veo3} & 56.3 &
52.1 & 50.8 & 43.0 & 53.0 & 50.4 &
63.4 & 61.0 & 68.9 & 63.7 \\

Seedance 1.0~\cite{gao2025seedance} & 55.1 &
54.2 & 42.5 & 44.8 & 45.4 & 44.2 &
62.2 & 64.1 & 69.8 & 68.6 \\

Kling 2.6 Pro~\cite{kling2025} & 53.4 &
52.9 & 59.8 & 36.4 & 53.0 & 35.8 &
57.0 & 60.5 & 63.7 & 61.3 \\

Sora v2 Pro~\cite{openai2025sora2} & 36.2 &
20.8 & 26.8 & 18.6 & 25.5 & 11.5 &
47.6 & 51.3 & 66.4 & 56.1 \\

Sora v1~\cite{openai2024sora} & 26.6 &
15.1 & 22.3 & 11.1 & 16.6 & 13.9 &
31.4 & 32.4 & 54.4 & 41.9 \\

\midrule
\rowcolor{RowBlue}
\multicolumn{11}{l}{\textbf{\textit{Robotics-specific}}} \\

Cosmos3-Super~\cite{nvidia2026worldsimulationvideofoundation} & 58.1 &
48.7 & 64.2 & 44.4 & 59.1 & 39.5 &
61.5 & 62.3 & \cellcolor{Top2}73.9 & \cellcolor{Top3}69.1 \\

Abot-PhysWorld~\cite{ali2025abot} & 52.9 &
48.6 & 54.8 & 43.4 & 52.3 & 45.4 &
66.2 & 66.8 & 53.1 & 45.5 \\

Cosmos 2.5~\cite{nvidia2026worldsimulationvideofoundation} & 46.4 &
35.8 & 33.8 & 20.1 & 49.6 & 39.9 &
54.4 & 56.0 & 65.8 & 62.6 \\

DreamGen(gr1)~\cite{jang2025dreamgen} & 42.0 &
31.2 & 37.2 & 29.7 & 33.4 & 21.5 &
56.4 & 53.2 & 57.9 & 57.5 \\

DreamGen(droid)~\cite{jang2025dreamgen} & 40.5 &
35.8 & 34.8 & 21.4 & 31.6 & 33.9 &
49.9 & 47.6 & 54.2 & 55.6 \\

Vidar~\cite{feng2025vidar} & 20.6 &
7.3 & 10.6 & 5.0 & 5.4 & 5.0 &
38.2 & 41.0 & 37.4 & 35.7 \\

UnifoLM-WMA-0~\cite{unifolm-wma-0} & 12.3 &
3.6 & 4.0 & 1.8 & 6.2 & 0.0 &
26.8 & 19.4 & 29.3 & 20.0 \\

\midrule
\rowcolor{RowBlue}
\multicolumn{11}{l}{\textbf{\textit{Finetuned methods}}} \\

Wan2.2-TI2V-5B~\cite{wan2025wan} & 38.0 &
33.1 & 31.3 & 14.2 & 31.8 & 23.4 &
43.6 & 44.8 & 59.0 & 60.7 \\

Wan2.2-TI2V-5B (ft) & 44.8 & 39.6 & 41.5 & 24.9 & 42.4 & 28.8 & 49.2 & 52.7 & 61.4 & 62.6  \\

\textbf{+ PhysisForcing} & 47.5 & 43.4 & 42.6 & 29.6 & 47.5 & 31.2 & 51.6 & 57.4 & 59.7 & 64.2 \\

\hline

Wan2.2-I2V-A14B~\cite{wan2025wan} & 50.7 &
38.1 & 45.4 & 37.3 & 50.1 & 33.0 &
60.8 & 58.2 & 69.0 & 64.8 \\

Wan2.2-I2V-A14B (ft) & 57.9 & 52.3 & 62.8 & 45.2 & 54.5 & 47.8 & 64.2 & 63.5 & 65.6 & 65.3  \\

\textbf{PF-Wan} & \cellcolor{Top2}62.0 & 56.4 & 65.4 & 49.1 & 58.4 & \cellcolor{Top3}52.4 & \cellcolor{Top2}68.7 & \cellcolor{Top2}69.6 & 69.2 & 68.5 \\

\hline

Cosmos3-Nano~\cite{nvidia2026worldsimulationvideofoundation} & 58.4 &
55.0 & \cellcolor{Top3}67.0 & 46.6 & 57.6 & 39.4 &
59.1 & 61.1 & \cellcolor{Top3}73.1 & 66.6 \\

Cosmos3-Nano (ft) & \cellcolor{Top3}61.5 & \cellcolor{Top2}57.8 & \cellcolor{Top2}67.4 & \cellcolor{Top3}49.2 & \cellcolor{Top2}59.3 & 48.5 & 66.5 & 67.1 & 70.6 & 67.1 \\

\textbf{PF-Cosmos} & \cellcolor{Top1}63.8 & \cellcolor{Top1}58.9 & \cellcolor{Top1}69.7 & \cellcolor{Top2}51.3 & \cellcolor{Top1}60.8 & \cellcolor{Top2}53.0 & \cellcolor{Top1}69.3 & \cellcolor{Top1}70.0 & 72.2 & 69.0 \\

\bottomrule
\end{tabular}

}
\label{tab:RBench_compare}
\end{table*}

\paragraph{R-Bench Evaluation.}
As shown in Table~\ref{tab:RBench_compare}, PhysisForcing improves every backbone.
PF-Cosmos attains the best overall score (63.8, $+9.2\%$ over base),
surpassing all baselines including the strongest commercial model Wan2.6 (60.7), while
PF-Wan reaches 62.0 ($+22.3\%$ over base), the second best overall, with consistent gains
holding on Wan2.2-TI2V-5B as well.

\paragraph{PAI-Bench Evaluation.}
\begin{figure}[t]
\centering
\begin{minipage}[t]{0.49\linewidth}
\centering
\includegraphics[width=\linewidth,keepaspectratio]{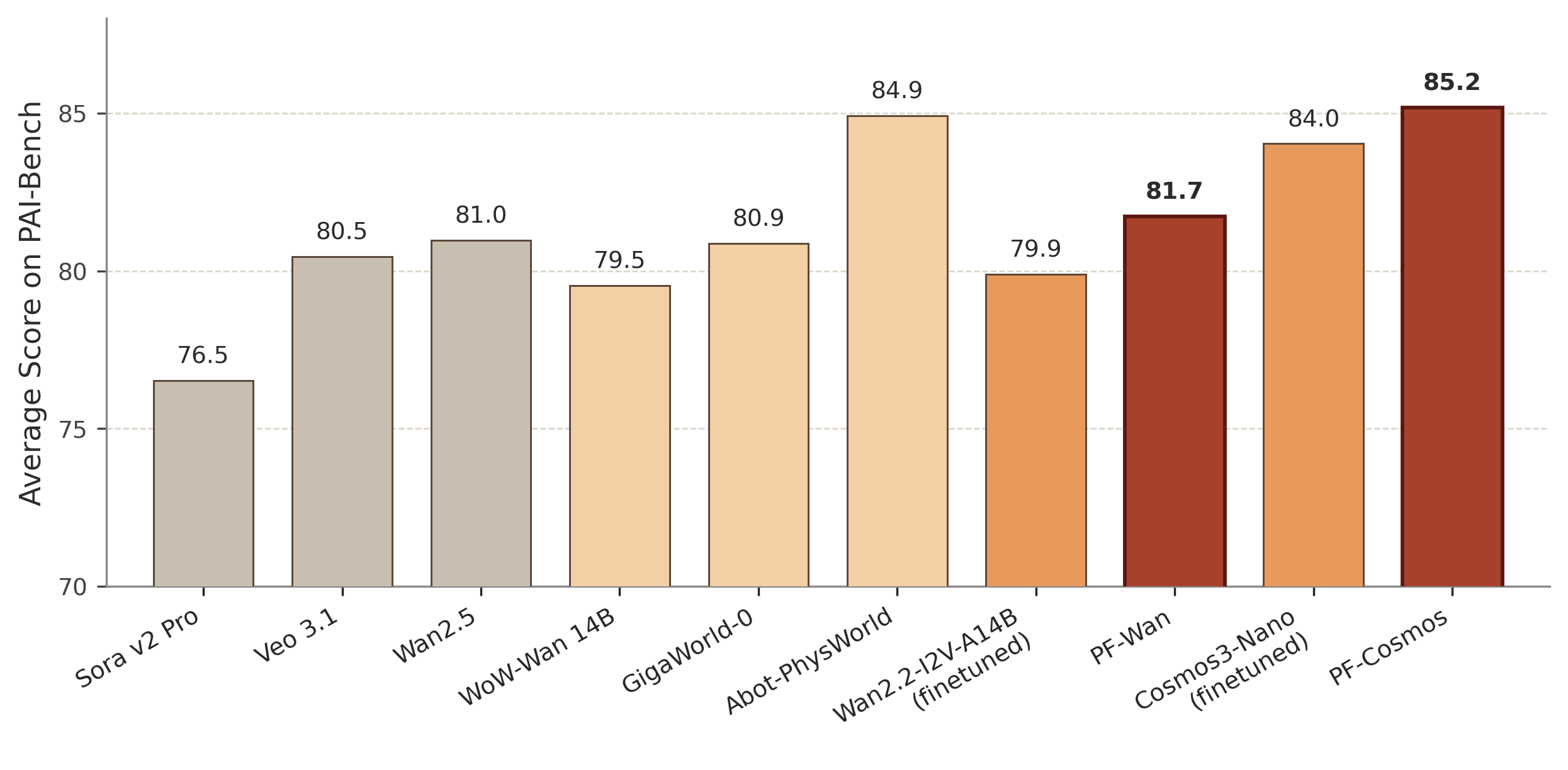}
\caption{Overall average score (mean of Quality and Domain scores) on the robot domain of PAI-Bench. Baseline scores are taken from the official leaderboard.}
\label{fig:paibench}
\end{minipage}
\hfill
\begin{minipage}[t]{0.49\linewidth}
\centering
\includegraphics[width=\linewidth,keepaspectratio]{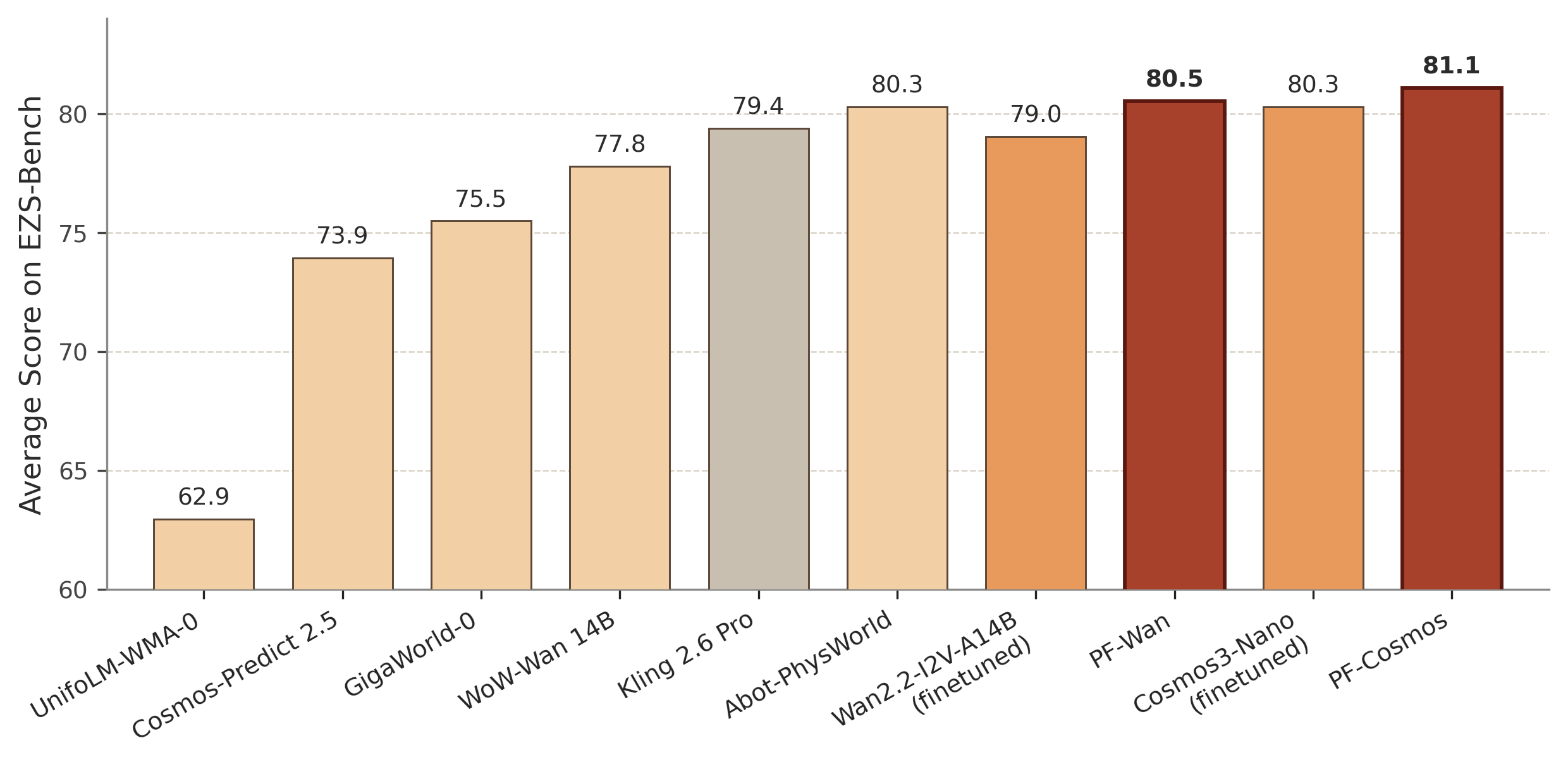}
\caption{Overall average score on EZS-Bench~\cite{ali2025abot}, a training-independent zero-shot benchmark of 196 unseen robot-task-scene combinations.}
\label{fig:ezsbench}
\end{minipage}
\end{figure}

On the robot domain of PAI-Bench (Figure~\ref{fig:paibench}), PhysisForcing improves both
backbones over vanilla finetuning (Wan2.2-I2V-A14B: $79.9\!\rightarrow\!81.7$;
Cosmos3-Nano: $84.0\!\rightarrow\!85.2$). PF-Cosmos attains the best
overall average (85.2), surpassing the strongest commercial model Wan2.5 (81.0) and
robotics-specific baseline Abot-PhysWorld (84.9). Full per-metric results are in
Appendix~\ref{appx:bench_results} (Table~\ref{tab:paibench_full}).

\paragraph{EZS-Bench Evaluation.}
On the training-independent zero-shot EZS-Bench~\cite{ali2025abot} (Figure~\ref{fig:ezsbench}),
which probes out-of-distribution generalization, PhysisForcing again improves both backbones
over vanilla finetuning (Wan2.2-I2V-A14B: $79.0\!\rightarrow\!80.5$;
Cosmos3-Nano: $80.3\!\rightarrow\!81.1$), with PF-Cosmos achieving the
best overall average (81.1), outperforming Abot-PhysWorld (80.3) and all other baselines.
Full per-metric results are in Appendix~\ref{appx:bench_results} (Table~\ref{tab:ezsbench_full}).

\begin{figure*}[!h]
\centering
\includegraphics[width=\textwidth]{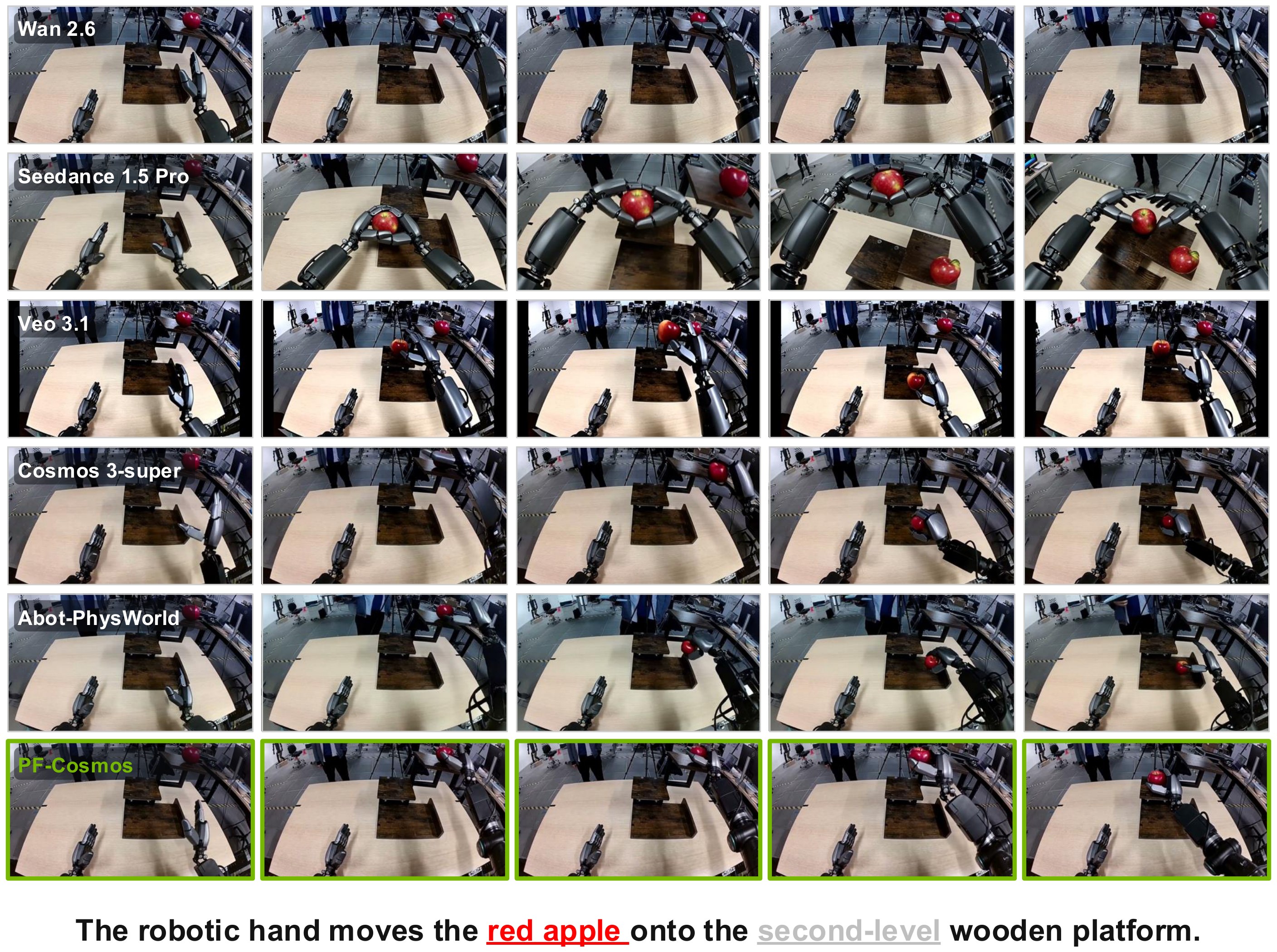}
\caption{\textbf{Qualitative comparison with state-of-the-art models.} Compared with strong commercial and robotics-specific models (Wan2.6, Seedance~1.5~Pro, Veo~3.1, Cosmos3-Super, Abot-PhysWorld), PhysisForcing (PF-Cosmos, last row in green) produces more physically plausible robot-object interactions. Prompt: moving the red apple onto the second-level wooden platform.}
\label{fig:qualitative}
\end{figure*}

\paragraph{Qualitative Analysis.}
As shown in Figure~\ref{fig:qualitative}, baselines often produce visually plausible but
physically inconsistent videos (e.g., unstable grasping, object drifting, incorrect contact),
whereas PhysisForcing better preserves robot-object contact and state transitions. More
qualitative results are in Appendix~\ref{appx:qualitative}.

\subsection{Evaluation results on Policy Learning}
\label{sec:wam}

We further evaluate PhysisForcing as the video backbone for world action modeling, plugging
the PhysisForcing-trained Wan2.2-TI2V-5B into Fast-WAM~\cite{yuan2026fastwam} as a drop-in
replacement for its video DiT on contact-rich RoboTwin~2.0 tasks~\cite{chen2025robotwin}.
We jointly train a single policy on the six tasks and evaluate each with 200 rollouts.
As shown in Table~\ref{tab:robotwin_results}, it improves the average success rate from
$68.2\%$ to $72.8\%$, with the largest gains on contact-rich placing and pressing
(\textit{place\_empty\_cup} $41.5\%\!\rightarrow\!63.0\%$, \textit{press\_stapler}
$49.0\%\!\rightarrow\!60.0\%$).

\paragraph{World model as an action planner.}
Under the action-planner protocol of WorldArena~\cite{shang2026worldarena}, the world model
is paired with a shared inverse dynamics model that decodes its predicted rollout into
actions executed in the RoboTwin~2.0 simulator. As shown in Table~\ref{tab:worldarena},
PhysisForcing lifts the average closed-loop success rate from $16.0\%$ to $24.0\%$,
surpassing all world-model planners including the strongest baseline WoW~\cite{chi2025wow}
($20.5\%$).

\begin{figure}[t]
\centering
\footnotesize
\setlength{\tabcolsep}{4pt}
\renewcommand{\arraystretch}{1.1}
\begin{minipage}[t]{0.49\linewidth}
\vspace{0pt}
\centering
\begin{minipage}[t][2.6\baselineskip][t]{\linewidth}
\captionof{table}{\textbf{Downstream policy success rate} (\%, 200 rollouts/task) on RoboTwin~2.0 (Fast-WAM backbone).}
\label{tab:robotwin_results}
\end{minipage}
\begin{tabular}{lccc}
\toprule
\textbf{Task} & \textbf{Baseline} & \textbf{PhysisForcing} & $\Delta$ \\
\midrule
\textit{place\_empty\_cup}  & 41.5 & \textbf{63.0} & \textcolor{teal}{$+21.5$} \\
\textit{press\_stapler}     & 49.0 & \textbf{60.0} & \textcolor{teal}{$+11.0$} \\
\textit{grab\_roller}       & 58.5 & \textbf{63.0} & \textcolor{teal}{$+4.5$} \\
\textit{shake\_bottle}      & \textbf{97.5} & 94.5 & $-3.0$ \\
\textit{adjust\_bottle}     & 93.0 & 93.0 & $\phantom{+}0.0$ \\
\textit{stack\_bowls\_two}  & \textbf{69.5} & 63.0 & $-6.5$ \\
\midrule
\textbf{Average}            & 68.2 & \textbf{72.8} & \textcolor{teal}{$+4.6$} \\
\bottomrule
\end{tabular}
\end{minipage}
\hfill
\begin{minipage}[t]{0.49\linewidth}
\vspace{0pt}
\centering
\begin{minipage}[t][2.6\baselineskip][t]{\linewidth}
\captionof{table}{\textbf{WorldArena action planner} closed-loop success rate (\%); best world model in \textbf{bold}.}
\label{tab:worldarena}
\end{minipage}
\begin{tabular}{lccc}
\toprule
\textbf{Model} & \textbf{Task 1} & \textbf{Task 2} & \textbf{Avg.} \\
\midrule
Genie Envisioner~\cite{liao2025genie} & 10.0 & 20.0 & 15.0 \\
TesserAct~\cite{zhen2025tesseract} & \phantom{0}1.0 & \textbf{35.0} & 18.0 \\
RoboMaster & \phantom{0}8.0 & 20.0 & 14.0 \\
Vidar~\cite{feng2025vidar} & \phantom{0}2.0 & 19.0 & 10.5 \\
WoW~\cite{chi2025wow} & 20.0 & 21.0 & 20.5 \\
\midrule
Wan2.2-TI2V-5B & 12.0 & 20.0 & 16.0 \\
\quad + \textbf{PhysisForcing} & \textbf{22.0} & 26.0 & \textbf{24.0} \\
\bottomrule
\end{tabular}
\end{minipage}
\end{figure}

\subsection{Ablation Study}
\label{sec:ablation}

\begin{figure}[t]
\centering
\footnotesize
\setlength{\tabcolsep}{4pt}
\renewcommand{\arraystretch}{1.0}
\begin{minipage}[t]{0.47\linewidth}
\vspace{0pt}
\centering
\captionof{table}{Per-component ablation on R-Bench.}
\label{tab:ablation_R-Bench}
\begin{tabular}{lccc}
\toprule
\textbf{Model} & \textbf{Emb.} & \textbf{Tasks} & \textbf{Avg.} \\
\midrule
Wan2.2-TI2V-5B (ft) & 56.5 & 35.4 & 44.8 \\
\quad + $\mathcal{L}^{\mathrm{phy}}_{\mathrm{pix}}$ & 59.0 & 37.8 & 47.2 \\
\quad + $\mathcal{L}^{\mathrm{phy}}_{\mathrm{sem}}$ & 58.4 & 36.5 & 46.2 \\
\quad + PhysisForcing & \textbf{58.2} & \textbf{38.9} & \textbf{47.5} \\
\midrule
Wan2.2-I2V-A14B (ft) & 64.7 & 52.5 & 57.9 \\
\quad + $\mathcal{L}^{\mathrm{phy}}_{\mathrm{pix}}$ & 67.5 & 55.2 & 60.7 \\
\quad + $\mathcal{L}^{\mathrm{phy}}_{\mathrm{sem}}$ & 66.8 & 54.6 & 60.0 \\
\quad + PhysisForcing & \textbf{69.0} & \textbf{56.3} & \textbf{62.0} \\
\bottomrule
\end{tabular}

\vspace{0.25em}

\captionof{table}{Physics region focus ablation on R-Bench.}
\label{tab:ablation_region}
\begin{tabular}{lccc}
\toprule
\textbf{Model} & \textbf{Emb.} & \textbf{Tasks} & \textbf{Avg.} \\
\midrule
Wan2.2-TI2V-5B (ft) & 56.5 & 35.4 & 44.8 \\
w/o Physics region focus & 57.0  & 37.2  & 46.0  \\
w/ Physics region focus & 58.2 & 38.9 & 47.5 \\
\bottomrule
\end{tabular}

\vspace{0.25em}

\captionof{table}{Alignment-block (layer index) ablation (Wan2.2-TI2V-5B, PAI-Bench robot domain).}
\label{tab:ablation_layer}
\begin{tabular}{l cccc}
\toprule
\textbf{Layer} & 10 & \textbf{15} & 20 & 25 \\
\midrule
\textbf{Robot Domain Score} & 83.9 & \textbf{85.2} & 84.1 & 83.2 \\
\bottomrule
\end{tabular}
\end{minipage}
\hfill
\begin{minipage}[t]{0.50\linewidth}
\vspace{0pt}
\centering
\includegraphics[width=\linewidth]{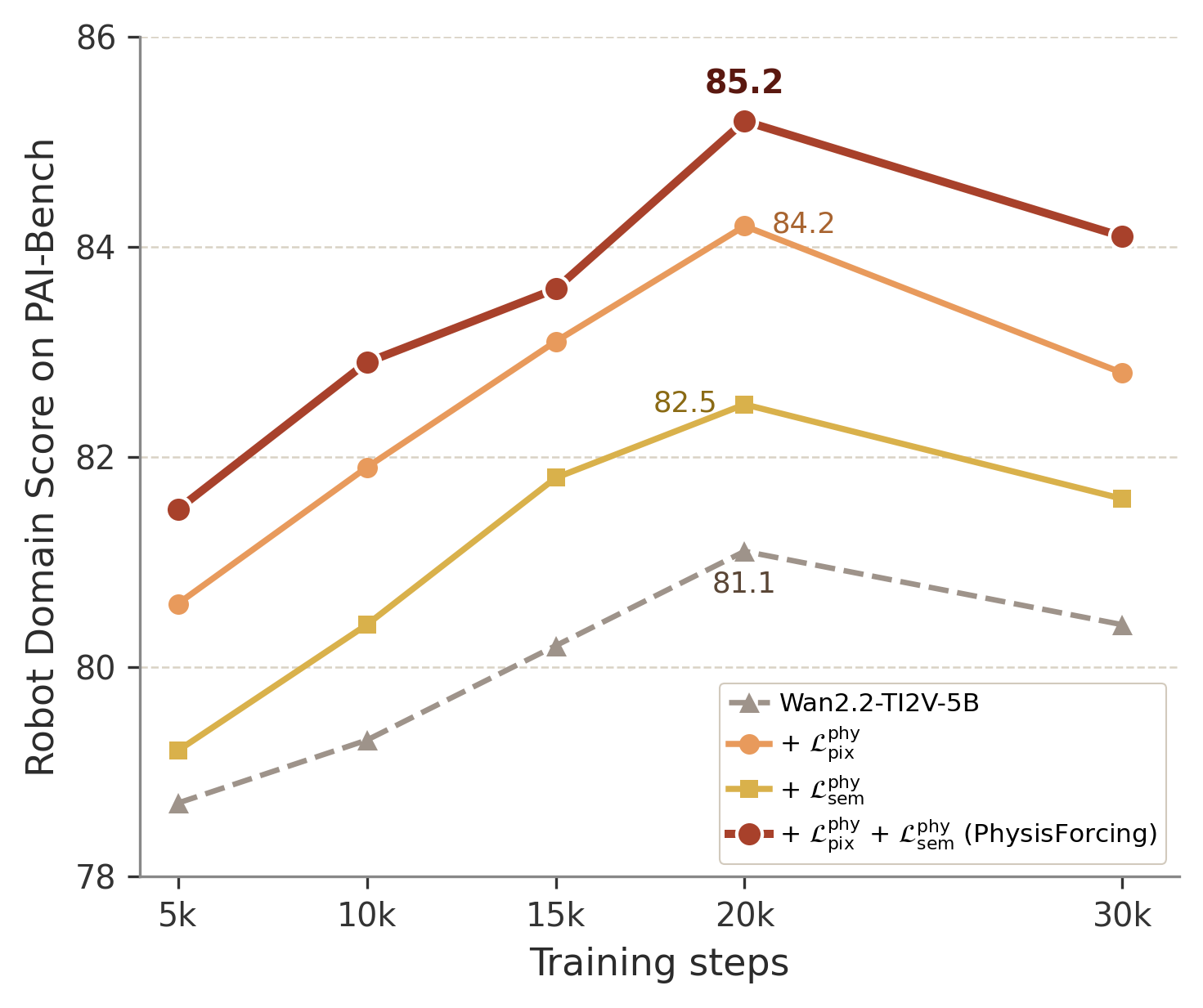}
\captionof{figure}{Robot domain score (PAI-Bench) across training steps. The two physics losses are complementary throughout training.}
\label{fig:training_curve}
\end{minipage}
\end{figure}

\paragraph{Component ablation.}
Table~\ref{tab:ablation_R-Bench} isolates each objective on R-Bench, and the two losses prove
complementary. On Wan2.2-TI2V-5B, the pixel-level trajectory loss
$\mathcal{L}^{\mathrm{phy}}_{\mathrm{pix}}$ and the semantic-level relational loss
$\mathcal{L}^{\mathrm{phy}}_{\mathrm{sem}}$ each improve the finetuned baseline ($44.8$) to
$47.2$ and $46.2$, and combining them is best ($47.5$).
$\mathcal{L}^{\mathrm{phy}}_{\mathrm{pix}}$ gives the larger single-loss gain because it
directly suppresses trajectory discontinuity, the most common local failure, whereas
$\mathcal{L}^{\mathrm{phy}}_{\mathrm{sem}}$ mainly repairs global relational errors such as
broken contact, so the two target different error modes and stack. The same pattern holds on
the larger Wan2.2-I2V-A14B ($57.9\!\rightarrow\!62.0$), confirming the benefit is not tied to
backbone scale.

\paragraph{Physics region focus.}
Table~\ref{tab:ablation_region} verifies the value of concentrating supervision on
interaction-critical regions. Applying the same two losses uniformly over all tokens already
helps ($44.8\!\rightarrow\!46.0$), but restricting them to physics-informative regions further
lifts the average to $47.5$, with the largest gain on the task-oriented dimensions
(\textit{Tasks} $35.4\!\rightarrow\!38.9$). This shows that background and near-static regions
dilute the physical signal, so focusing where robot-object interactions occur is what drives
task-level correctness.

\paragraph{Training dynamics.}
Figure~\ref{fig:training_curve} tracks the PAI-Bench robot domain score over training. Both
losses outperform vanilla finetuning at every checkpoint and the full model leads throughout,
indicating a persistent rather than transient learning signal. The score peaks at $85.2$ at
20k ($+4.1$); all variants then decline slightly from mild overfitting, yet PhysisForcing
still leads by $+3.7$ at 30k.

\paragraph{Alignment layer.}
Because the alignment is imposed on a single DiT block, its depth matters. Sweeping the block
index on Wan2.2-TI2V-5B (Table~\ref{tab:ablation_layer}), a middle block (layer $15$) is best
($85.2$), beating shallower (layer $10$, $83.9$) and deeper (layer $25$, $83.2$) choices: early
blocks still carry mostly shallow appearance features and lack the semantic structure needed
for relational alignment, while late blocks are already specialized for the final noise prediction and
are harder to steer. The intermediate block thus offers the best trade-off, and performance
stays stable across nearby layers.
\section{Conclusion}

We present PhysisForcing, a training-time framework that improves robotic video generation by
aligning pixel-level trajectories and semantic-level relations on interaction-critical regions.
Across R-Bench, PAI-Bench, and the zero-shot EZS-Bench, it consistently surpasses base models,
vanilla finetuning, and strong open-source, commercial, and robotics-specific baselines, with
PF-Cosmos best overall on all three. It further raises the WorldArena closed-loop success rate
from $16.0\%$ to $24.0\%$ and improves downstream policy success, showing that physical
plausibility yields concrete benefits for embodied intelligence.

\clearpage

\bibliographystyle{plainnat}
\setlength{\bibhang}{0pt}
\setlength\bibindent{0pt}
\bibliography{main}

\clearpage
\appendix

{
  \centering 
  \textbf{\Large PhysisForcing: Physics Reinforced World Simulator for Robotic Manipulation}
  
  \vspace{5pt}
  {\Large Appendix}
  \par
}

\section{More implementation details}
\label{appx:impl}

This section complements the main paper with the concrete configuration of the
auxiliary perception models that produce our physics targets, and of the three video
diffusion backbones that we fine-tune with PhysisForcing.

\paragraph{Auxiliary perception models.}
All three auxiliary models are frozen and used only to extract physics targets;
they are run on the ground-truth clip on the fly during each training step, and
the tracker/depth outputs are shared between the two physics losses. For the
semantic-level teacher $\mathcal{E}_\phi$ in $\mathcal{L}^{\mathrm{phy}}_{\mathrm{sem}}$
we adopt \textbf{V-JEPA\,2}~\cite{bardes2024revisiting} in its public
\emph{ViT-L/16} variant (the \texttt{vitl-fpc64-256} checkpoint, hidden width
$1024$), a self-supervised video encoder trained by predicting masked
spatio-temporal feature targets rather than pixels. Its tokens are known to
capture object- and interaction-centric structure, which makes its token
relation matrix a natural semantic-level target. The clip is mapped to $[0,1]$,
trilinearly resampled to $64{\times}256{\times}256$ and ImageNet-normalized, and
the encoder returns a $32{\times}16{\times}16$ spatio-temporal token grid
(tubelet $2$, patch $16$). Taking the implementation on Wan2.2-I2V-A14B as a reference, we take the
hidden feature of a single DiT block (block $20$, width $5120$), map it into
V-JEPA\,2's feature space with a lightweight MLP, and trilinearly resample it from its native DiT patch grid to
the same $32{\times}16{\times}16$ grid, so that student and teacher tokens are
index-aligned; the semantic-level loss then compares the $K{\times}K$ pairwise-cosine
relation matrices over the (up to $K{=}512$) mask-selected tokens, rather than
the absolute features. For the reference trajectories in
$\mathcal{L}^{\mathrm{phy}}_{\mathrm{pix}}$ we use \textbf{CoTracker3}~\cite{karaev24cotracker3}
in its \emph{offline} variant, a transformer point tracker with factorized
spatio-temporal attention. We initialize a regular $25{\times}25$ query grid
($625$ points) on the first frame and run it on the same clip fed to the
diffusion model, yielding per-point 2D trajectories and visibilities. To favor
foreground motion under scene-scale ambiguity, we additionally run
\textbf{Depth-Anything-V2}~\cite{yang2024depth} (the ViT-L variant,
$\sim$335M parameters) as a frozen monocular depth estimator \emph{on the first
frame}; the resulting relative depth map, normalized to $[0,1]$, weights each
track's motion magnitude so that the top-$K$ most active foreground tracks are
retained. These selected trajectories serve both as the pixel-level supervision
and, after rasterization, as the physics-informative mask $M$ used by the
semantic-level loss.

\paragraph{Wan2.2-I2V-A14B backbone.}
The first backbone is \textbf{Wan2.2-I2V-A14B}~\cite{wan2025wan}, a Mixture-of-Experts image-to-video
diffusion transformer with two $\sim$14B-parameter denoiser experts (27B total, 14B
active per step) operating on top of the original Wan2.1 3D causal VAE with a
$T{\times}H{\times}W = 4{\times}8{\times}8$ spatio-temporal compression ratio. The two
experts are split along the diffusion trajectory by an SNR-based boundary $t_{\mathrm{moe}}$:
the \emph{high-noise} expert covers timesteps $t \geq t_{\mathrm{moe}}$ and is responsible
for laying down global layout, motion, and object configuration, while the
\emph{low-noise} expert covers $t < t_{\mathrm{moe}}$ and refines high-frequency
appearance. PhysisForcing targets the dynamics-forming stage, where physical structure
is committed; we therefore fine-tune only the \emph{high-noise} expert. This is also
well-aligned with our data setup: our robot manipulation training videos are at
$480{\rm p}$, where directly fine-tuning the high-noise expert is sufficient to
specialize the model to the target distribution. We deviate
from the original MoE routing during training and apply the high-noise expert across the
full $t \in [0, T]$ range: each training step samples $t$ uniformly from the entire
diffusion schedule and feeds the noisy latent through the high-noise expert regardless of
whether $t$ falls above or below $t_{\mathrm{moe}}$. The two physics losses
$\mathcal{L}^{\mathrm{phy}}_{\mathrm{pix}}$ and $\mathcal{L}^{\mathrm{phy}}_{\mathrm{sem}}$
are added to the standard flow-matching objective on these jointly sampled timesteps, so
the fine-tuned expert learns to obey the trajectory and relational constraints uniformly
across the denoising trajectory while the low-noise expert is left untouched.

\paragraph{Wan2.2-TI2V-5B backbone.}
Our smaller backbone is \textbf{Wan2.2-TI2V-5B}~\cite{wan2025wan}, a unified text/image-to-video diffusion
transformer with a single $\sim$5B-parameter denoiser. Unlike the A14B model, TI2V-5B is
paired with the new \textbf{Wan2.2-VAE}, whose $T{\times}H{\times}W = 4{\times}16{\times}16$
compression ratio yields an overall $64{\times}$ reduction of the input volume; together
with the patchify layer of the diffusion transformer, the effective compression seen by
the denoiser becomes $4{\times}32{\times}32$. Because TI2V-5B is a single-expert model, no MoE routing is needed: we
directly fine-tune the full denoiser with the standard flow-matching loss together with
$\mathcal{L}^{\mathrm{phy}}_{\mathrm{pix}}$ and $\mathcal{L}^{\mathrm{phy}}_{\mathrm{sem}}$,
with $t$ sampled uniformly across the full diffusion schedule. All other training
hyper-parameters are kept identical between the two backbones to isolate the effect of the backbone itself.

\paragraph{Cosmos3-Nano backbone.}
The third backbone is \textbf{Cosmos3-Nano}~\cite{nvidia2026worldsimulationvideofoundation}, a $\sim$16B-parameter
\emph{Mixture-of-Transformers} (MoT) video model built on the Qwen3-VL-8B
backbone (hidden width $4096$, $36$ transformer blocks). We operate
the model in its image-to-video setting and fine-tune it on the same
robot-manipulation clips at $720{\rm p}$. The latent space is produced by the same
\textbf{Wan2.2-VAE} as Wan2.2-TI2V-5B ($T{\times}H{\times}W = 4{\times}16{\times}16$),
which together with the diffusion transformer's patchify layer yields an effective
$4{\times}32{\times}32$ compression seen by the denoiser. Following the same recipe as the Wan backbones, we read the hidden
feature of a single mid-depth MoT block and add the two physics losses to conduct the physics supervision.

\section{Evaluation benchmark details}
\label{appx:benchmarks}

We provide further details on the embodied video generation benchmarks used in the main
text: R-Bench~\cite{deng2026rethinking}, the robot domain of
PAI-Bench~\cite{zhou2025paibench}, and EZS-Bench~\cite{ali2025abot}.

\paragraph{R-Bench.}
R-Bench consists of 650 image-text prompt pairs that cover diverse robotic manipulation and
locomotion scenarios, with each pair annotated along two orthogonal axes. The
\emph{task} axis contains five categories, namely Manipulation, Spatial Relationship,
Multi-Entity Collaboration, Long-Horizon Planning, and Visual Reasoning, ranging from
single-step pick-and-place to multi-agent and reasoning-heavy scenarios. The
\emph{embodiment} axis contains four categories, namely Single-Arm, Dual-Arm, Quadruped,
and Humanoid robots, testing whether the model can consistently render robot morphologies
and their motion patterns. The per-dimension scores reported in the main text correspond
exactly to these nine sub-categories, and the overall score is their unweighted average.
For each generated video, R-Bench computes five fine-grained sub-metrics that jointly
cover task-level correctness and visual fidelity. \emph{Physical-Semantic Plausibility}
and \emph{Task-Adherence Consistency} are evaluated through an MLLM-as-Judge VQA pipeline:
uniformly sampled key frames are arranged into temporal grids and inspected by Qwen3-VL,
which is prompted to detect physical violations such as floating components, object
penetration, spontaneous appearance/disappearance, and non-contact attachment, and to
verify that prompt-specified key actions occur in the correct order. \emph{Robot-Subject
Stability} uses a contrastive VQA protocol that compares a reference frame against a
generated frame to detect gripper or object morphology drift, link-topology changes, and
attribute drift over time. \emph{Motion Smoothness} and \emph{Motion Amplitude} are
computed from pixel- and tracking-based statistics rather than from MLLM prompting,
penalizing temporal jitter as well as degenerate near-static outputs. The five sub-metrics
are normalized and aggregated into a single per-prompt score in $[0, 1]$. The benchmark
has been validated to correlate strongly with human judgement, reporting a Spearman rank
correlation of $0.96$ between R-Bench scores and crowd-sourced preferences across
$25$ representative video generators. We use the official evaluation scripts and report
results on all 650 prompts.

\paragraph{PAI-Bench (robot domain).}
PAI-Bench is a comprehensive Physical AI benchmark spanning six domains (autonomous
vehicles, robotics, industry, common sense, human, physics) and three tracks (PAI-Bench-G
for video generation, PAI-Bench-C for conditional generation, PAI-Bench-U for video
understanding). In this paper we evaluate exclusively on the \emph{robot subset of the
PAI-Bench-G generation track}, which is the subset most directly aligned with our embodied
manipulation setting. It contains 174 image-prompt pairs sourced from real-world robotic
captures, each accompanied by 5-6 QA pairs that encode the expected physical and
semantic content. Each prompt is produced by a two-stage curation pipeline in which an
MLLM (Qwen2.5-VL-72B-Instruct) generates an initial caption that is then manually
refined, and domain-grounded QA pairs are generated and manually curated from the same
source video. In this work we report both the \emph{Domain Score} and the \emph{Quality Score} of
PAI-Bench-G on the robot subset. The Domain Score directly measures the physical and
semantic plausibility of generated robotic interactions: it adopts the MLLM-as-Judge
paradigm with Qwen3-VL-235B-A22B-Instruct, where each curated QA pair is posed to the MLLM
against the generated video and the score is the MLLM's response accuracy across the QA
set. The benchmark validates this metric with an arena-based human study and reports an
overall Pearson correlation of $0.918$ with human ELO ratings. The Quality Score
aggregates the benchmark's multidimensional perceptual metrics (subject/background
consistency, overall consistency, aesthetic and imaging quality, and motion smoothness).
The overall average reported in Figure~\ref{fig:paibench} and Table~\ref{tab:paibench_full}
is the mean of the Quality and Domain scores, averaged over the 174 robot prompts of
PAI-Bench-G.

\paragraph{EZS-Bench.}
EZS-Bench~\cite{ali2025abot} is a training-independent embodied zero-shot benchmark
designed to evaluate physical fidelity and cross-embodiment generalization under fully
out-of-distribution conditions, where diverse robot morphologies, environments, and tasks
are composed into previously unseen combinations with no overlap with training data. The
evaluation set is built with a dual-branch strategy: one branch synthesizes initial frames
with a text-to-image model by varying robots, scenes, tasks, and viewpoints, while the
other applies VLM-guided background editing to real robot images while preserving the
foreground interaction. Each initial image is paired with a physics-grounded dense
description that integrates the initial state, action trajectory, and final state, yielding
196 evaluation samples. EZS-Bench adopts a decoupled dual-model protocol that separately
scores visual quality and physical-semantic plausibility: a Qwen3-VL-32B-Thinking model
generates a physical checklist from the initial state and instruction, and a separate
Qwen2.5-VL-72B-Instruct model answers the checklist against the generated video, which
avoids the self-evaluation bias of using a single model as both question generator and
judge. Following the PAI-Bench convention, we report the Quality Score, the Domain Score,
and their average as the overall score.

\section{Detailed benchmark results}
\label{appx:bench_results}

We report the Quality Score and Domain Score behind the average scores shown in
Figure~\ref{fig:paibench} and Figure~\ref{fig:ezsbench} of the main text in more detail.
Table~\ref{tab:paibench_full} reports the full PAI-Bench (robot domain) results and
Table~\ref{tab:ezsbench_full} reports the full EZS-Bench results. On both benchmarks, PhysisForcing improves the Domain Score and the
overall average of both backbones over their vanilla-finetuned counterparts, and
PF-Cosmos achieves the best overall average, surpassing the
strongest robotics-specific baseline Abot-PhysWorld~\cite{ali2025abot}. The gains are most
pronounced on the Domain Score, which measures physical-semantic plausibility, while the
Quality Score stays competitive, indicating that hierarchical physics alignment improves
physical fidelity while keeping visual quality on par with vanilla finetuning.

\begin{table}[t]
\centering
\renewcommand{\arraystretch}{1.15}
\setlength{\tabcolsep}{10pt}
\caption{\textbf{Full results on the robot domain of PAI-Bench.} We report the Quality Score, the Domain Score, and their average (overall Avg.). \textbf{(ft)} denotes vanilla finetuning of the corresponding backbone. The best value in each column is in \textbf{bold}.}
\label{tab:paibench_full}
\begin{tabular}{l ccc}
\toprule
\textbf{Model} & \textbf{Quality} & \textbf{Domain} & \textbf{Avg.} \\
\midrule
Wan2.5~\cite{wan2025wan}            & 75.48 & 86.44 & 80.96 \\
GigaWorld-0~\cite{team2025gigaworld} & 75.91 & 85.83 & 80.87 \\
Veo 3.1~\cite{GoogleDeepMind2025Veo3} & \textbf{77.40} & 83.50 & 80.45 \\
WoW-Wan 14B~\cite{chi2025wow}        & 76.05 & 83.01 & 79.53 \\
Sora v2 Pro~\cite{openai2025sora2}   & 76.79 & 76.26 & 76.52 \\
Abot-PhysWorld~\cite{ali2025abot}    & 76.76 & 93.06 & 84.91 \\
Wan2.2-I2V-A14B~\cite{wan2025wan}        & 76.15 & 81.70 & 78.93 \\
\midrule
Wan2.2-I2V-A14B (ft)~\cite{wan2025wan}   & 75.38 & 84.42 & 79.90 \\
\quad \textbf{PF-Wan}                & 76.26 & 88.20 & 81.73 \\
Cosmos3-Nano (ft)                    & 76.52 & 91.54 & 84.03 \\
\quad \textbf{PF-Cosmos}             & 77.08 & \textbf{93.26} & \textbf{85.17} \\
\bottomrule
\end{tabular}
\end{table}

\begin{table}[t]
\centering
\renewcommand{\arraystretch}{1.15}
\setlength{\tabcolsep}{10pt}
\caption{\textbf{Full results on EZS-Bench.} EZS-Bench~\cite{ali2025abot} is a training-independent embodied zero-shot benchmark of 196 unseen robot-task-scene combinations. We report the Quality Score, the Domain Score, and their average (overall Avg.). \textbf{(ft)} denotes vanilla finetuning of the corresponding backbone. The best value in each column is in \textbf{bold}.}
\label{tab:ezsbench_full}
\begin{tabular}{l ccc}
\toprule
\textbf{Model} & \textbf{Quality} & \textbf{Domain} & \textbf{Avg.} \\
\midrule
WoW-Wan 14B~\cite{chi2025wow}        & 76.09 & 79.51 & 77.80 \\
GigaWorld-0~\cite{team2025gigaworld} & 72.72 & 78.26 & 75.49 \\
Cosmos-Predict 2.5~\cite{ali2025world} & 70.89 & 76.98 & 73.94 \\
UnifoLM-WMA-0~\cite{unifolm-wma-0}   & 73.55 & 52.32 & 62.94 \\
Kling 2.6 Pro~\cite{kling2025}       & \textbf{78.05} & 80.72 & 79.39 \\
Abot-PhysWorld~\cite{ali2025abot}    & 76.94 & 83.66 & 80.30 \\
Wan2.2-I2V-A14B~\cite{wan2025wan}        & 76.89 & 77.42 & 77.16 \\
\midrule
Wan2.2-I2V-A14B (ft)~\cite{wan2025wan}   & 76.12 & 81.95 & 79.04 \\
\quad \textbf{PF-Wan}                & 76.58 & 84.49 & 80.54 \\
Cosmos3-Nano (ft)                    & 77.42 & 83.16 & 80.29 \\
\quad \textbf{PF-Cosmos}             & 76.95 & \textbf{85.20} & \textbf{81.08} \\
\bottomrule
\end{tabular}
\end{table}

\section{More qualitative results}
\label{appx:qualitative}

\begin{figure}[h]
\centering
\includegraphics[width=\linewidth]{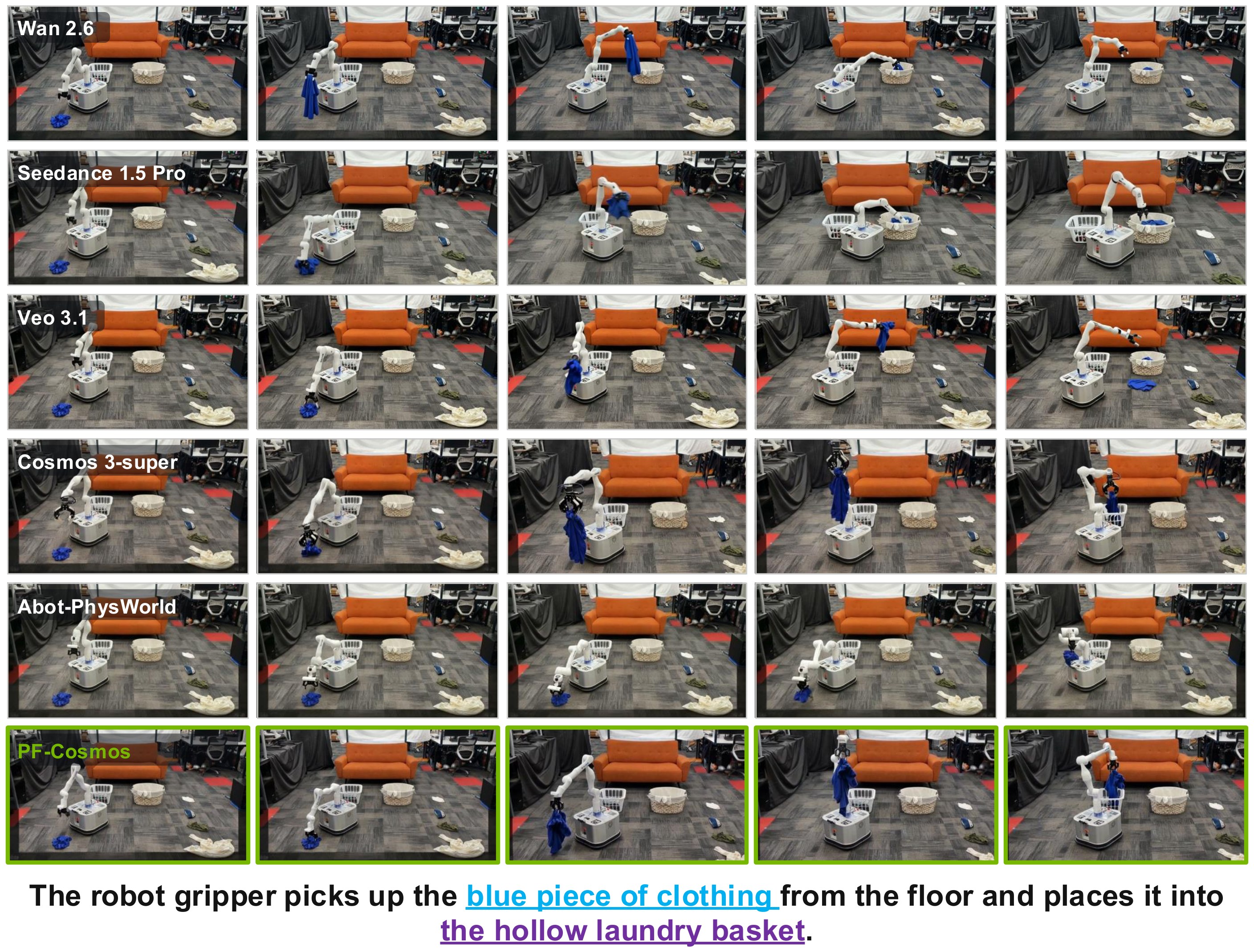}
\caption{Comparison with the state-of-the-art models.}
\label{fig:appx_qual_sota1}
\end{figure}

\begin{figure}[h]
\centering
\includegraphics[width=\linewidth]{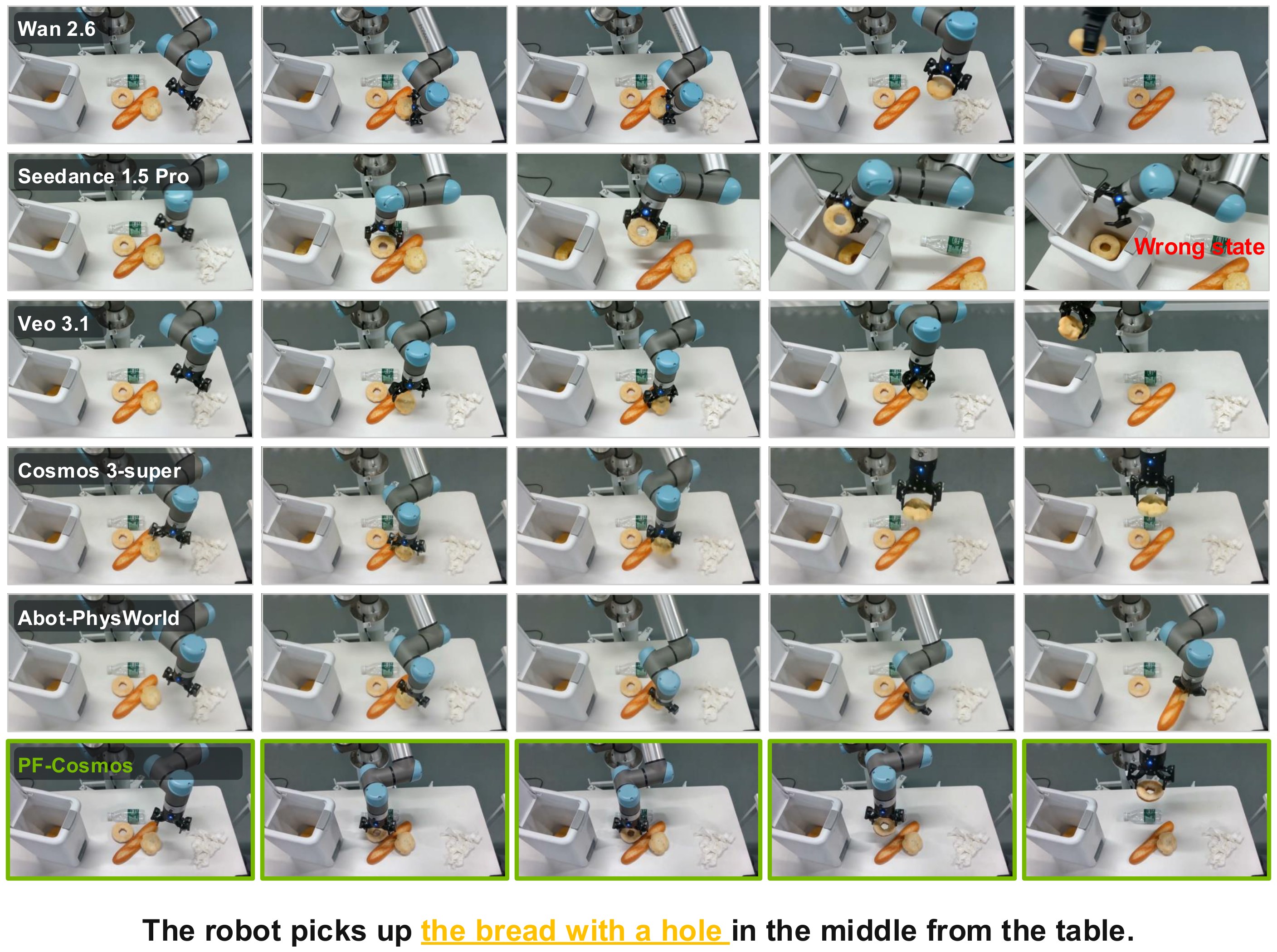}
\caption{Comparison with the state-of-the-art models.}
\label{fig:appx_qual_sota2}
\end{figure}

\begin{figure}[h]
\centering
\includegraphics[width=\linewidth]{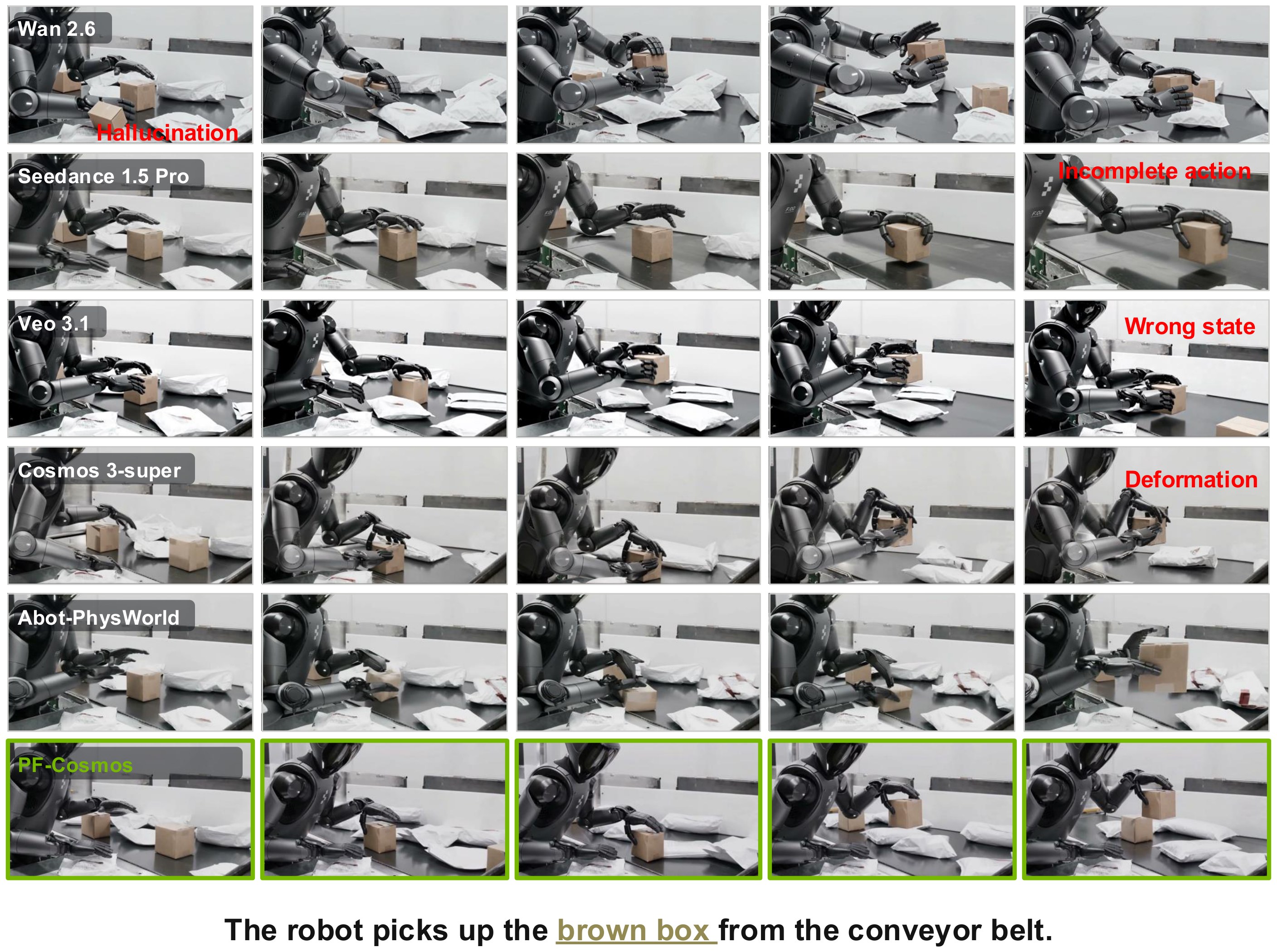}
\caption{Comparison with the state-of-the-art models.}
\label{fig:appx_qual_sota3}
\end{figure}

\begin{figure}[h]
\centering
\includegraphics[width=\linewidth]{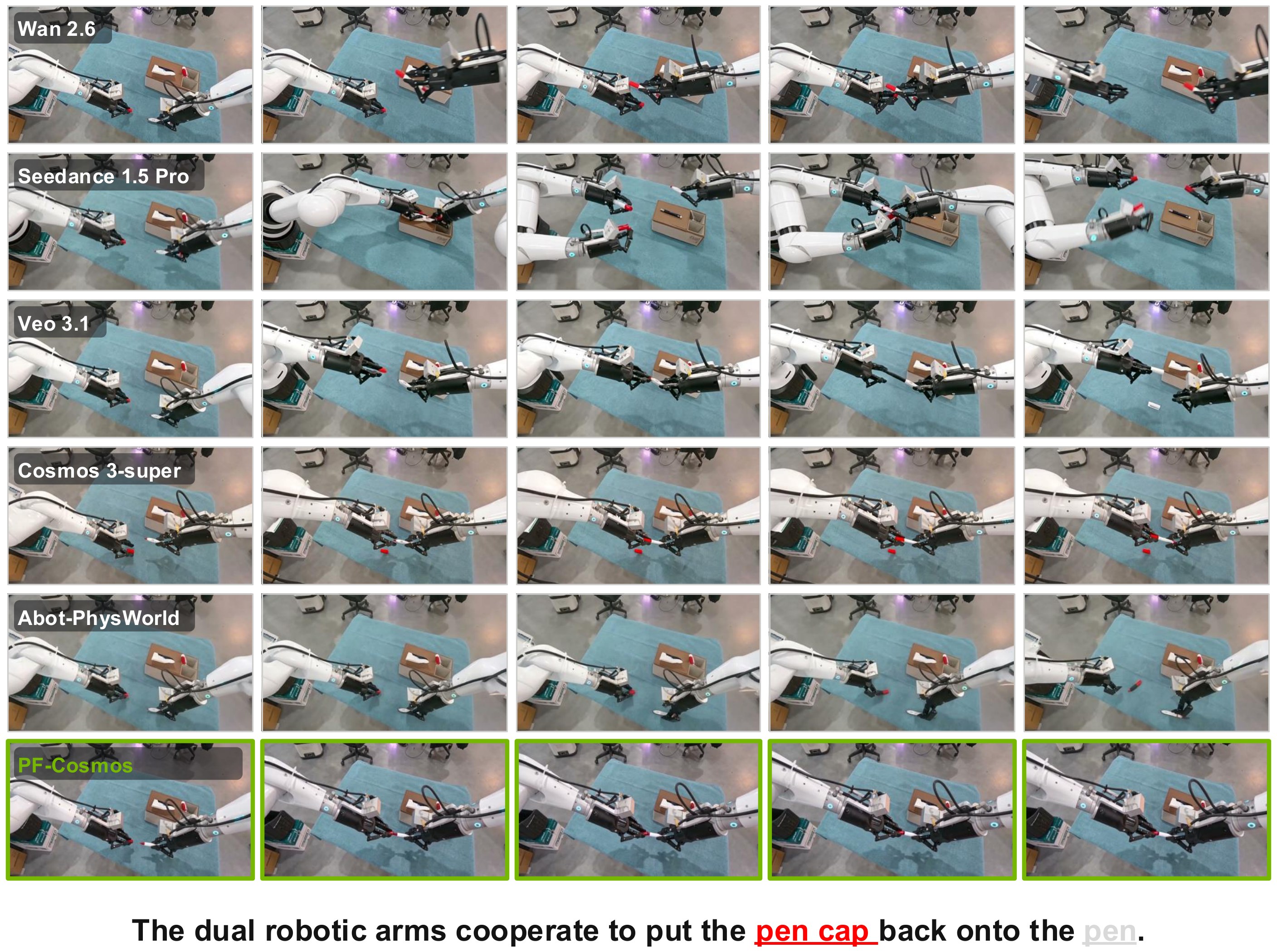}
\caption{Comparison with the state-of-the-art models.}
\label{fig:appx_qual_sota4}
\end{figure}

\begin{figure}[h]
\centering
\includegraphics[width=\linewidth]{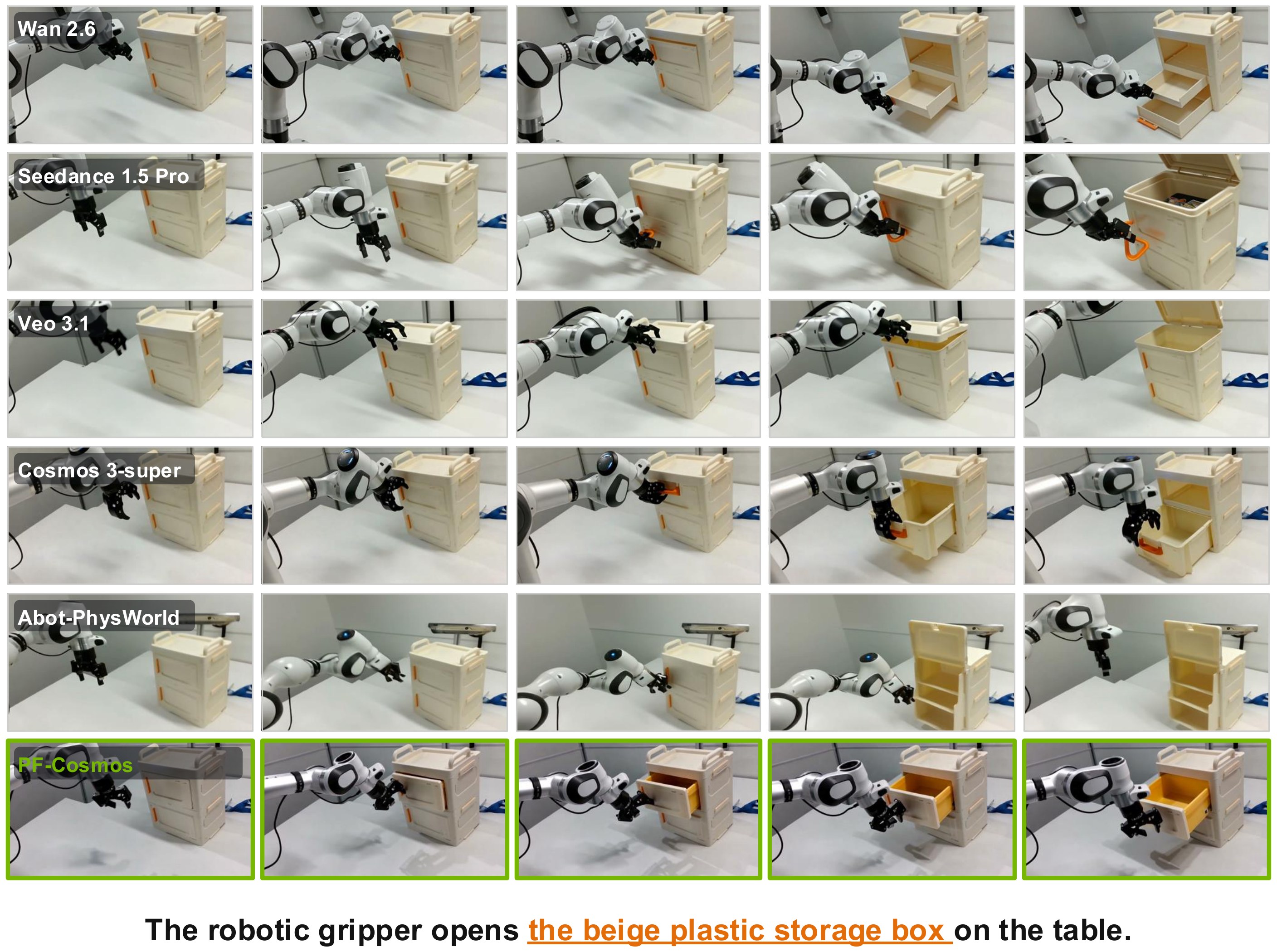}
\caption{Comparison with the state-of-the-art models.}
\label{fig:appx_qual_sota5}
\end{figure}

\paragraph{Comparison with the state-of-the-art models.}
Figures~\ref{fig:appx_qual_sota1}--\ref{fig:appx_qual_sota5} compare PhysisForcing
(PF-Cosmos) with strong commercial and robotics-specific models---Wan2.6, Seedance~1.5~Pro,
Veo~3.1, Cosmos3-Super, and Abot-PhysWorld---on shared prompts. Across diverse single-arm,
dual-arm, and humanoid tasks, the competing models frequently drift to wrong states, break
robot-object contact, or deform objects, while PhysisForcing produces more continuous motion
and more consistent interactions.

\begin{figure}[h]
\centering
\includegraphics[width=\linewidth]{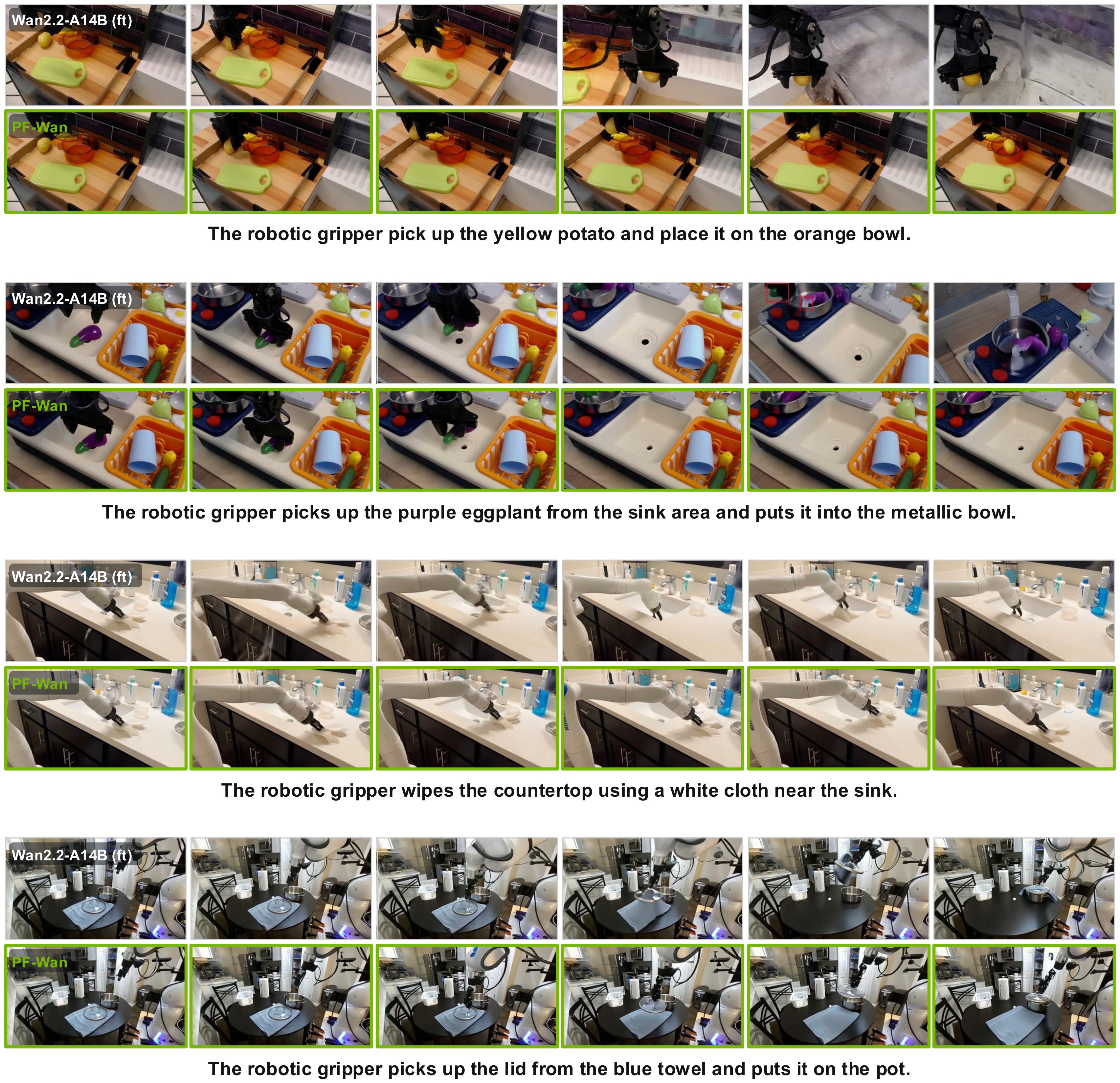}
\caption{Effect of PhysisForcing on the Wan2.2-I2V-A14B backbone: finetuned baseline (top) vs.
PF-Wan (bottom, green).}
\label{fig:appx_ablation_wan}
\end{figure}

\begin{figure}[h]
\centering
\includegraphics[width=\linewidth]{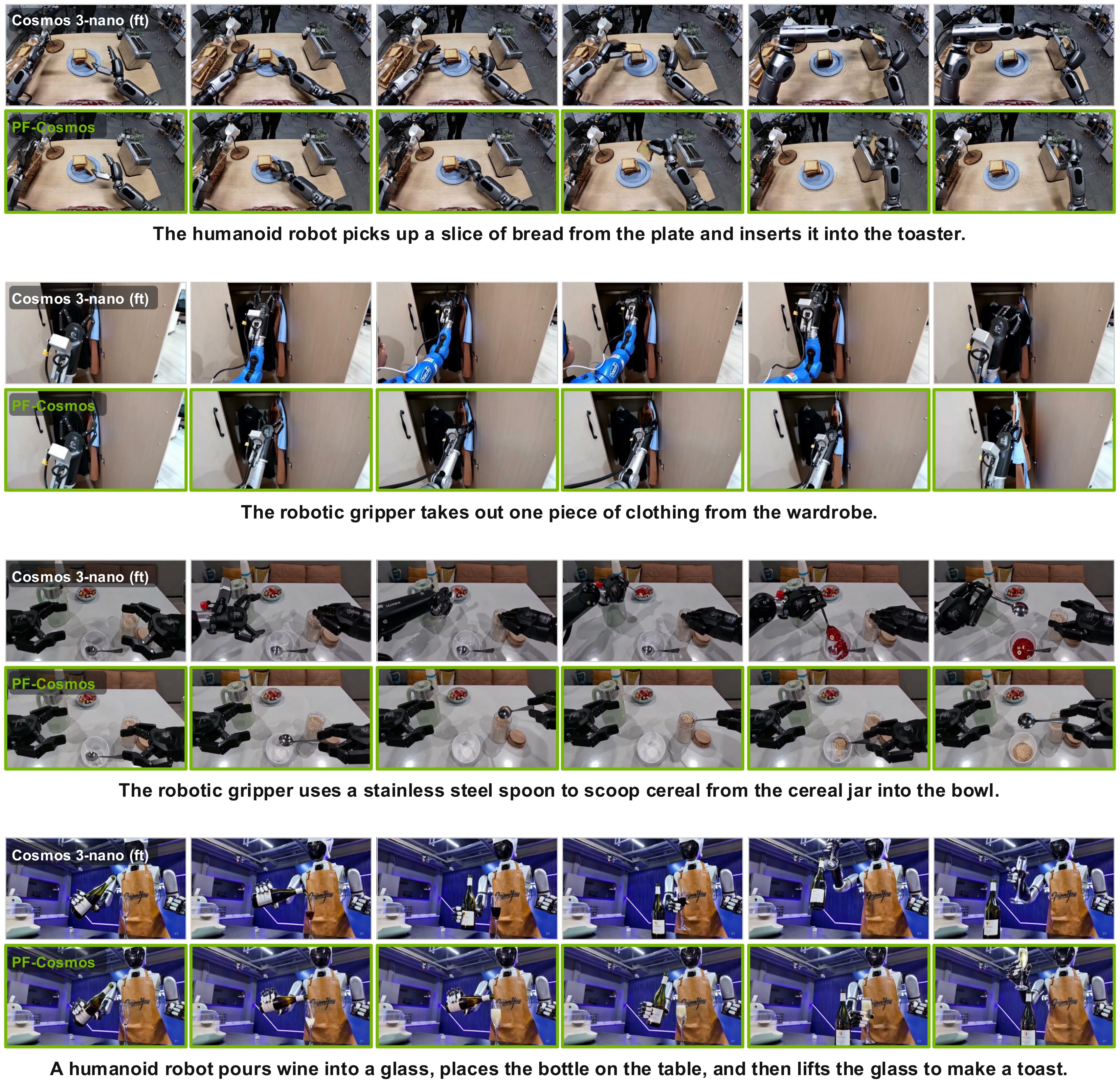}
\caption{Effect of PhysisForcing on the Cosmos3-Nano backbone: finetuned baseline (top) vs.
PF-Cosmos (bottom, green).}
\label{fig:appx_ablation_cosmos}
\end{figure}

\paragraph{Improving physical plausibility with PhysisForcing.}
Figures~\ref{fig:appx_ablation_wan} and~\ref{fig:appx_ablation_cosmos} present qualitative
ablations that isolate the effect of PhysisForcing from vanilla finetuning on the two backbones.
For each prompt, the top row is the finetuned baseline and the bottom row (green) is the same
backbone trained with PhysisForcing. On Wan2.2-I2V-A14B (Figure~\ref{fig:appx_ablation_wan}), the
finetuned baseline often deforms the manipulated object or loses robot-object contact, whereas
PF-Wan keeps object shape stable and produces continuous grasp-and-place trajectories. The same
trend holds on Cosmos3-Nano (Figure~\ref{fig:appx_ablation_cosmos}), where PF-Cosmos better
maintains contact dynamics and inter-object relations across single-arm, dual-arm, and humanoid
tasks. These comparisons show that the gains stem from the hierarchical physics alignment rather
than from additional in-domain finetuning.

\begin{figure}[h]
\centering
\includegraphics[width=\linewidth]{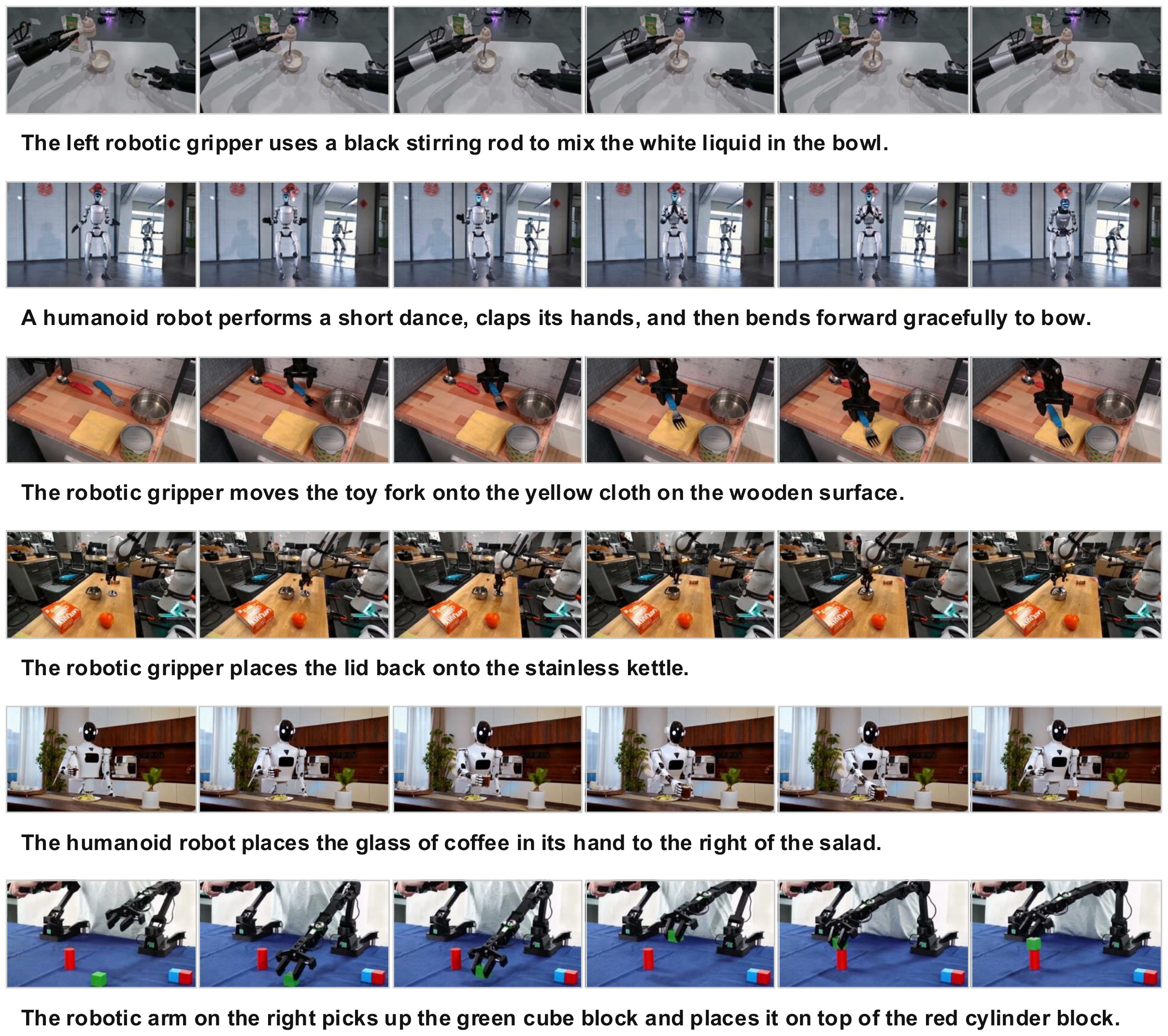}
\caption{More generation results of PhysisForcing (PF-Cosmos).}
\label{fig:appx_qual_gen1}
\end{figure}

\begin{figure}[h]
\centering
\includegraphics[width=\linewidth]{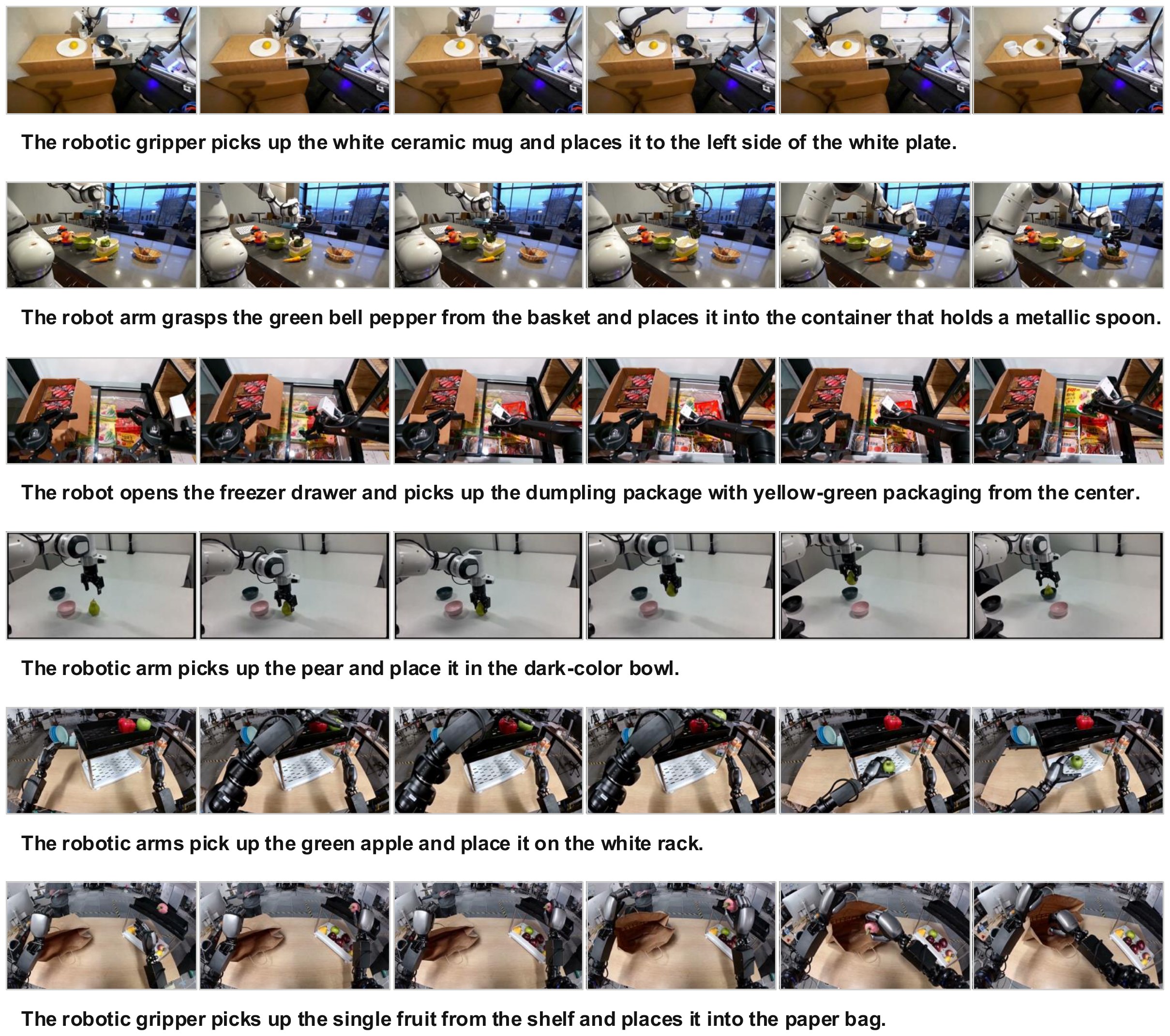}
\caption{More generation results of PhysisForcing (PF-Cosmos).}
\label{fig:appx_qual_gen2}
\end{figure}

\paragraph{Generalization across scenes and instructions.}
Figures~\ref{fig:appx_qual_gen1} and~\ref{fig:appx_qual_gen2} show additional generation
results produced by PhysisForcing (PF-Cosmos) across diverse embodiments, scenes, and
instructions. PhysisForcing consistently follows the prompt while respecting basic physical
regularities, indicating that the learned physical priors generalize beyond the comparison
prompts above.

\section{Broader impacts}
\label{appx:impact}

PhysisForcing improves the physical plausibility of generated robotic manipulation
videos. On the positive side, this lowers the barrier to embodied AI research by
providing a cheap simulator for data augmentation and pre-deployment policy evaluation,
and yields more reliable training signals for downstream world models. On the negative
side, more realistic robot videos may be misused to fabricate deceptive footage,
overstate hardware capabilities, or seed policies trained on synthetic demonstrations
without real-world grounding. We mitigate these risks via a research-only release, and
recommend that downstream users combine the model with provenance tools and validate
any synthetic-data-derived policy on real hardware before deployment.

\section{Limitations and future work}
\label{appx:limitations}

PhysisForcing is a fine-tuning recipe and inherits the capability ceiling of its
underlying backbone: current open-source video generators, including the Wan2.2 and Cosmos3 families
we build on, still exhibit limited world knowledge and long-horizon temporal reasoning,
which bounds the physical plausibility any fine-tuning method can reach. As stronger
video and world-model foundation models become available, we expect PhysisForcing
to compound with their capabilities, since its trajectory-level and relational physics
constraints are model-agnostic and only grow more informative as the backbone's
representational capacity increases.

\end{document}